\documentclass[twoside,11pt]{article}

\usepackage{jmlr2e}
\usepackage{lineno,hyperref}
\modulolinenumbers[5]
\usepackage[utf8]{inputenc}
\usepackage[T1]{fontenc}
\usepackage{graphicx}
\usepackage{grffile}
\usepackage{longtable}
\usepackage{wrapfig}
\usepackage{rotating}
\usepackage[normalem]{ulem}
\usepackage{amsmath}
\DeclareMathOperator{\arcsinh}{arcsinh}

\usepackage{textcomp}
\usepackage{amssymb}
\usepackage{capt-of}
\usepackage{hyperref}
\usepackage{mathtools}
\usepackage{centernot}
\usepackage{subfig}
\usepackage{booktabs}
\hypersetup{colorlinks=true}

\newcommand{\Intr}{\operatorname{int}({\sgens{k}_+})}

\newcommand{\real}{\mathbb{R}}
\newcommand\df{\stackrel{\mathclap{\tiny\mbox{df.}}}{=}}
\newcommand{\sgens}[1]{\mathbb{R}^{#1}}
\newcommand{\sgen}[2]{\mathbb{R}^{#1\times #2}}

\renewcommand{\cal}[1]{\mathcal{#1}}
\newcommand{\op}[1]{\operatorname{#1}}

\newcommand{\Fix}[1]{\operatorname{Fix}({#1})}

\newcommand{\NA}{---}
\newcommand{\closure}[2][3]{%
{}\mkern#1mu\overline{\mkern-#1mu#2}}
\usepackage{enumitem}
\newcommand{\subscript}[2]{$#1 _ #2$}
\makeatletter
\newcommand{\proofpart}[2]{%
  \par
  \addvspace{\medskipamount}%
  \noindent\emph{Part #1: #2}\par\nobreak
  \addvspace{\smallskipamount}%
  \@afterheading
}
\makeatother
\newtheorem{assumption}[theorem]{Assumption}
\newtheorem{fact}[theorem]{Fact}

\usepackage{lastpage}
\jmlrheading{25}{2024}{1-\pageref{LastPage}}{2/23; Revised
3/24}{5/24}{23-0167}{Tomasz J. Piotrowski, Renato L. G. Cavalcante, Mateusz Gabor}
\ShortHeadings{title}{Piotrowski, Cavalcante, Gabor}

\firstpageno{1}

\begin{document}

\title{Fixed points of nonnegative neural networks}

\author{\name Tomasz J. Piotrowski \email tpiotrowski@umk.pl \\
       \addr Faculty of Physics, Astronomy and Informatics,\\
       Nicolaus Copernicus University,\\
       Grudziądzka 5/7, 87-100 Toruń, Poland
       \AND
       \name Renato L. G. Cavalcante \email renato.cavalcante@hhi.fraunhofer.de\\
       \addr Fraunhofer Heinrich Hertz Institute, \\
       Einsteinufer 37,  10587 Berlin, Germany
       \AND
       \name Mateusz Gabor \email mateusz.gabor@pwr.edu.pl\\
       \addr Faculty of Electronics, Photonics, and Microsystems, \\
       Wrocław University of Science and Technology,\\
       Wybrzeze Wyspianskiego 27, 50-370 Wrocław, Poland}

\editor{Aryeh Kontorovich}

\maketitle

\begin{abstract}
We use fixed point theory to analyze nonnegative neural networks, which we define as neural networks that map nonnegative vectors to nonnegative vectors. We first show that nonnegative neural networks with nonnegative weights and biases can be recognized as monotonic and (weakly) scalable mappings within the framework of nonlinear Perron-Frobenius theory. This fact enables us to provide conditions for the existence of fixed points of nonnegative neural networks having inputs and outputs of the same dimension, and these conditions are weaker than those recently obtained  using arguments in convex analysis. Furthermore, we prove that the shape of the fixed point set of nonnegative neural networks with nonnegative weights and biases is an interval, which under mild conditions degenerates to a point. These results are then used to obtain the existence of fixed points of more general nonnegative neural networks. From a practical perspective, our results contribute to the understanding of the behavior of autoencoders, and we also offer valuable mathematical machinery for future developments in deep equilibrium models. 
\end{abstract}


\section{Introduction} \label{intro}
Neural networks consist of multiple layers that are able to extract patterns of input signals for decision-making without any human intervention. This fact has profound consequences in a wide range of wireless communications, image recognition, and speech processing tasks, where neural networks set high standards of
performance \cite{LeCun2015,Goodfellow2016,Samek2021}. Neural networks have also  been recently introduced as efficient iterative regularization methods in
inverse problems, which opens a vista of new applications, especially in medical imaging \cite{Adler2017,Benning2018,Ongie2020}.

Our study is specifically focused on nonnegative neural networks, defined as neural networks that map nonnegative vectors to nonnegative vectors. These networks have been widely used in many applications such as spectral analysis, image processing, and text processing  \cite{lemme2012online,nguyen2013learning,chorowski2015learning,hosseini2015deep,zhou2016,ali2017automatic,ayinde2018deep,su2018stacked,chen2019,yu2019,palsson2019spectral}. Nonnegative networks can be designed to decompose input signals into additive sparse components, leading to the formation of an interpretable hierarchical representation of the input data, mirroring the way the human brain processes sensory inputs \cite{Lee1999,Shashanka2008,lemme2012online,nguyen2013learning,chorowski2015learning,ayinde2018deep}. This principle has been exploited in autoencoders with nonnegative weights, which aims to break down input data into additive features within the encoding layer, which are then additively recombined in the decoding layer to reconstruct the original data  \cite{lemme2012online,ayinde2018deep}. In some scenarios, the gains in transparency and interpretation obtained by restricting the weights of the neural network to be nonnegative do not translate into a significant loss of predictive power \cite{nguyen2013learning,chorowski2015learning}.

Neural networks with nonnegative weights have also been increasingly gaining attention in recent years because of the resurgence of analog neural networks, which have been demonstrated to be both readily implementable and more energy efficient than their digital counterparts \cite{kendall2020training,Paliy2021}. In analog networks, the weight parameters are physically encoded in the conductance of programmable resistors. Thus, these analog neural networks are examples of neural networks that are nonnegative because of the laws of physics. Moreover, various approaches can be used to mitigate the loss of predictive power caused by the physical nonnegative constraints \cite{kendall2020training}. These results are consistent with those obtained with the digital neural networks in \cite{nguyen2013learning,chorowski2015learning}.

With the objective of improving our understanding of nonnegative neural networks, we shed light on questions related to the existence of fixed points and the shape of the fixed point set of nonnegative neural networks (with matching input and output spaces). To illustrate the importance of fixed point analysis using concrete examples, let us first consider autoencoders, which can be seen as self-mappings taking the form $T:\real^k\to\real^k$ with $x\mapsto T_1(T_2(x))$, where $T_1:\sgens{k}\to\sgens{l}$ with $k>l$ is a mapping (encoder) used to learn compact representations of typical inputs, and $T_2:\sgens{l}\to\sgens{k}$ is a mapping (decoder) used to reconstruct the input from the compact representation obtained with $T_1$. From this mathematical perspective, the fixed points of an autoencoder $T$ are exactly the inputs that can be perfectly reconstructed. It is therefore important to determine whether the fixed point set of an autoencoder is nonempty, and, if so, how rich this set is (i.e., determine its shape).

As a second example of the significance of fixed point analysis in neural networks, we mention deep equilibrium (DEQ) models \cite{bai2019deep,Bai2020,Gilton2021,Huang2021,Tsuchida2023}, which have been increasingly gaining attention from the machine learning community in recent years. The core concept of implicit models is that, instead of designing layers explicitly, as usually done, it suffices to define a layer's objective (target) functional equation to hold, without actually specifying how such a layer should achieve this goal. More precisely, a DEQ layer maps an input $x\in\sgens{k}$ to an output $z^\star \in \textup{Fix}(g_x)\subset\sgens{k}$, where $g_x:\sgens{k}\to\sgens{k}$ is an explicit mapping parameterized by $x.$ In the above definition, closed-form expressions for the implicit function $x\mapsto z^\star \in \textup{Fix}(g_x)$ are not required, and the numerical scheme used to compute the output $z^\star \in \textup{Fix}(g_x)$ from a given input $x$ is not specified. As a result, to ensure that the implicit function $x\mapsto z^\star \in \textup{Fix}(g_x)$ is well-defined and independent of the numerical scheme used to compute the output, it is natural to require $\textup{Fix}(g_x)$ to be a singleton for every $x\subset X$. This fixed point formulation also allows for direct implicit differentiation, which is crucial for the efficiency of training of DEQ models \cite{bai2019deep}. 

Fixed point analysis in neural networks is also important in the context of inverse problems, which are often formulated as optimization problems with the objective of minimizing a cost function $f\colon\mathbb{R}^k\to \mathbb{R}$. These problems are typically solved with algorithms that can be interpreted as fixed point iterations of mappings determined by $f$. For example, to minimize a cost function $f\colon\mathbb{R}^k\to \mathbb{R}$, we can use the standard gradient descent method: $x_{n+1}=(Id-\alpha\nabla f)(x_n),\ n=0,1,2,\dots$. The objective of this algorithm is to locate a point $x^\star$ satisfying $\nabla f(x^*)=0$, which is also a fixed-point of the mapping $(Id-\alpha\nabla f)$, where $Id\colon\sgens{k}\to\sgens{k}\colon x\mapsto x$ is the identity mapping. Many other iterative algorithms for optimization have a similar interpretation: they can be seen as fixed point algorithms based on a particular instance of the Mann iteration \cite{yamada11}. One potential limitation of this traditional mathematical approach for inverse problems is that the function $f$ is often imperfectly known (e.g., owing to the uncertainty about the noise distribution), so, in machine learning, data is used to train a neural network that is used as the mapping of a Mann iteration. In this approach, known as \emph{loop unrolling}, the number of layers is directly related to the number of fixed point iterations. As a result, these neural networks effectively emulate a fixed-point seeking algorithm. 



In light of the above discussion, it is clear that studying the properties of the fixed point set of neural networks is crucial for the development of general theory able to explain their behavior in many applications. To this end, we show that, for nonnegative neural networks with nonnegative weights and biases, one of the tractable paths for establishing the existence and the shape of fixed point sets is the nonlinear Perron-Frobenius (PF) theory \cite{Lemmens2012}, which is used to study monotonic and weakly scalable mappings defined on a cone (e.g., the nonnegative orthant). As shown in this manuscript, nonnegative neural networks with nonnegative weights and biases can be considered as instances of these mappings under mild and natural assumptions, which opens up the possibility for establishing the existence of fixed points of general nonnegative neural networks. In more detail, we organize the discussion of our results as follows:
\begin{enumerate}
\item In Section \ref{kc} we introduce notation and key concepts.  
\item Section \ref{wsinn} contains the key technical contributions of our study:
\begin{itemize}
\item We first show that nonnegative neural networks with nonnegative weights and biases are, under mild assumptions, both monotonic and (weakly) scalable in the sense of Definition \ref{main_df} introduced later. This result is a major departure from a recent line of research \cite{chen2019,Combettes2020a,Combettes2020b} that has used convex analysis in Hilbert spaces to study the behavior of neural networks.\footnote{In the framework of nonlinear PF theory, monotonicity and (weak) scalability have been widely exploited in the context of wireless networks \cite{boche2008,martin11,shindoh2019,shindoh2020,cavalcante2015elementary,Cavalcante2019,Cavalcante2019_IEEE_TSP,renato14SPM,renato2016maxmin,you2020note}, economic studies \cite{Oshime1992,woodford2022}, and biological systems \cite{krause2015positive}, to cite a few fields.}
\item Using the monotonicity and weak scalability properties of nonnegative neural networks with nonnegative weights and biases, we then proceed to characterize the shape of the fixed point set of these neural networks in the interior of the nonnegative cone $\sgens{k}_+$. In particular, we prove that the shape of the fixed point set is often an interval, which degenerates to a point if the neural network is not only weakly scalable, but also scalable. We show that these conditions are weaker than those found in some previous studies, which typically use arguments based on the nonexpansivity of activation functions in Hilbert spaces and operator norms of weight operators \cite{Combettes2020a,Combettes2020b}.
\end{itemize}  
\item In Section \ref{wsinn_arb}, we give conditions that guarantee that general monotonic nonnegative neural networks have at least one fixed point, and we show that the standard fixed point iteration converges to a fixed point.
\item In Section \ref{renato_ext}, we provide conditions for the existence of fixed point(s) of general nonnegative neural networks in settings where Brouwer's fixed point theorem is not directly applicable. 
\item In Section \ref{app}, we illustrate the main theoretical results via numerical simulations, where we evaluate the reconstruction performance of various nonnegative autoencoders. We also discuss how the theoretical contributions of this study provide us with a powerful mathematical machinery for designing efficient DEQ models. Results in this direction are already available in \cite{Gabor2024}.
\item In Section \ref{frd}, we outline and discuss future research directions, which build on both the theoretical and numerical results provided in Sections \ref{wsinn}-\ref{app}.
\end{enumerate}

\section{Key concepts} \label{kc}
We start by introducing key concepts and definitions pertaining to vectors and mappings on the nonnegative cone~$\sgens{k}_+.$ Let $u,v\in\sgens{k}_+$; then, $u\leq v$ denotes the partial ordering induced by the nonnegative cone, i.e., $u \le v \Leftrightarrow v-u\in\sgens{k}_+$. Similarly, $u<v$ means that $v-u\in\sgens{k}_{+}$ and $u\neq v$, while $u\ll v$ means that $v-u\in\op{int}(\sgens{k}_+)$, where $\op{int}(\sgens{k}_+)$ denotes the interior of the nonnegative cone $\sgens{k}_+.$

\begin{definition} \label{main_df}
A continuous mapping $f\colon\sgens{s}_+\to\sgens{p}$ is said to be
\begin{enumerate} [label=(\subscript{A}{{\arabic*}}), start=0]
       \item \emph{nonnegative} if
       \begin{equation} 
       \forall x\in\sgens{s}_+\quad f(x)\in\sgens{p}_+,
       \end{equation}
       \item \emph{monotonic} if
       \begin{equation} 
       \forall x,\tilde{x}\in\sgens{s}_+\quad x\leq\tilde{x}\implies f(x)\leq f(\tilde{x}),
       \end{equation}
       \item \emph{weakly scalable} if
       \begin{equation}
       \forall x\in\sgens{s}_+\quad \forall \rho\geq 1\quad f(\rho x)\leq \rho f(x),
       \end{equation}
        \item \emph{scalable} if
       \begin{equation} 
       \forall x\in\sgens{s}_+\quad \forall \rho> 1\quad f(\rho x)\ll\rho f(x).
       \end{equation}       
 \end{enumerate}
We use the convention that each subscript applied to $A$ refers to one of the above properties, so that, for example: 
\begin{itemize}
\item continuous and nonnegative mappings are $(A_0)$-mappings;
\item continuous, nonnegative, and monotonic mappings are $(A_{0,1})$-mappings;
\item continuous, nonnegative, monotonic, and weakly scalable mappings are $(A_{0,1,2})$-mappings; and
\item continuous, nonnegative, monotonic, and scalable mappings are $(A_{0,1,2,3})$-mappings.
\end{itemize}
\end{definition}

The above classes of mappings satisfy the relations $A_{0,1,2,3}\subset A_{0,1,2}\subset A_{0,1}\subset A_{0}$  by definition. We will also use the class $A_{0,2}$ of continuous, nonnegative, and weakly scalable mappings in conjunction with a stricter than $A_1$ notion of monotonicity.
\begin{definition} \label{strict_strong}
An $(A_0)$-mapping $f\colon\sgens{s}_+\to\sgens{p}_+$ is said to be \emph{strictly monotonic} if
\begin{equation}
\forall x,\tilde{x}\in\sgens{s}_+\quad x<\tilde{x}\implies f(x)<f(\tilde{x})\quad \text{and}\quad
x\ll\tilde{x}\implies f(x)\ll f(\tilde{x}).
\end{equation}
\end{definition}

We now introduce the standard definition of a neural network.

\begin{definition} \label{nn}
Let $T_i\colon\sgens{k_{i-1}}\to\sgens{k_i}$ of the form $T_i(x_{i-1})\df\sigma_i(W_ix_{i-1}+b_i)$ be the $i$-th layer of an $n$-layered feed forward neural network, $i=1,\dots, n$, where $x_{i-1}\in\sgens{k_{i-1}}$ is the input to the layer,  $W_i\colon\sgens{k_{i-1}}\to\sgens{k_i}$ is the linear weight operator (matrix), $b_i\in\sgens{k_i}$ is the bias, and $\sigma_i\colon\sgens{k_i}\to\sgens{k_i}$ is the activation function. A neural network $T$ is then the composition
\begin{equation} \label{dnn}
  T\df T_n\circ\dots\circ T_1,
\end{equation}  
with the zeroth layer corresponding to the input to the network. Hereafter, we assume that the input and output layers have the same dimension $k\df k_0=k_n$, which enables us to consider traditional recurrent neural networks and autoencoders as particular cases.
\end{definition}

We also assume that the layers of a neural network $T$ taking the form in (\ref{dnn}) use separable activation functions in the sense defined below.
\begin{definition} \label{sep_activ}
Let $T_i$ be the $i$-th layer of a neural network $T$ in (\ref{dnn}). An activation function $\sigma_i\colon\sgens{k_i}\to\sgens{k_i}$ for this layer is composed of $k_i$ scalar activation functions $\sigma_{i_j}\colon\real\to\real$ ($j=1,2,\dots,k_i$), and its operation is defined as 
\begin{equation} \label{activ_layer}
\sigma_i(y_i)=\sum_{j=1}^{k_i}(\sigma_{i_j}(y_{i_j}))e_{i_j},
\end{equation}
where $y_{i_j}\in\mathbb{R}$ is the $j$-th coefficient of the vector $y_i\df W_ix_{i-1}+b_i\in\sgens{k_i}$ and $e_{i_j}$ is the $j$-th coefficient of the canonical basis of $\,\sgens{k_i}$ for $j=1,2,\dots,k_i.$ For later reference, we equivalently express a layer $T_i$ with the activation function $\sigma_i$ in (\ref{activ_layer}) as
\begin{equation} \label{whew}
T_i(x_{i-1})=\sigma_i(y_i)=\Big(\sigma_{i_1}(y_{i_1}),\sigma_{i_2}(y_{i_2}),\dots,\sigma_{i_{k_i}}(y_{i_{k_i}})\Big),
\end{equation}
where $\sigma_{i_j}$ are scalar activation functions for $j=1,2,\dots,k_i.$
\end{definition}

We are now ready to study the set of fixed points of nonnegative neural networks, which are networks that have already been used in the literature and that are increasingly gaining attention because of the new trends in analog computation \cite{nguyen2013learning,lemme2012online,ayinde2018deep,chorowski2015learning,hosseini2015deep,chen2019,ali2017automatic,palsson2019spectral,su2018stacked,kendall2020training,Paliy2021}. With the theory we develop in Sections \ref{wsinn} and \ref{wsinn_arb}, we can, for example, understand some limitations of autoenconders and gain insights into useful approaches for improving their reconstruction performance.

\section{Fixed points of $(A_{0,1,2})$-neural networks} \label{wsinn}
We begin with the following assumption.

\begin{assumption} \label{A012}
In Section \ref{wsinn} we assume that, for all layers $i=1,\dots,n$:
\begin{enumerate}
\item the weight matrices $W_i$ and biases $b_i$ have nonnegative coefficients; and
\item the activation functions $\sigma_i$ are $(A_{0,1,2})$-mappings.
\end{enumerate}
\end{assumption}

In Corollary \ref{ex1} below, we show that Assumption \ref{A012} guarantees that the resulting network is an $(A_{0,1,2})$-neural network, and, in certain cases, even an $(A_{0,1,2,3})$-neural network. However, we first demonstrate that Assumption \ref{A012}.2 holds for activation functions constructed with commonly used scalar activation functions, and we show an explicit construction of neural networks satisfying Assumption \ref{A012}.

\subsection{Construction of $(A_{0,1,2})$-activation functions}
Many commonly used scalar activation functions ($\sigma_{i_j}$ in Definition~\ref{sep_activ}) are continuous and concave in the nonnegative orthant. Therefore, in light of [Lemma 1]\cite{Cavalcante2019}, they belong to a proper subclass of $(A_{0,1,2})$-scalar activation functions. Some of them are listed below. 

\begin{remark} \label{activ}
The following two lists provide examples of widely-used continuous scalar concave activation functions (with their domains restricted to $\xi\in\mathbb{R}_+$), and, hence,  $(A_{0,1,2})$-scalar activation functions. We divide them into two classes:
\begin{itemize}
\item  functions satisfying $\lim_{\xi\to\infty}\sigma'(\xi)=0$, and
\item functions satisfying $\lim_{\xi\to\infty}\sigma'(\xi)=1.$
\end{itemize}
\begin{enumerate}
\item[(L1)] continuous scalar concave activation functions satisfying $\lim_{\xi\to\infty}\sigma'(\xi)=0$:
  \begin{itemize}
  \item (sigmoid)
    $\xi\mapsto \frac{1}{1+\exp{(-\xi)}}$
  \item (capped ReLU)
    $\xi\mapsto \min\{\xi,\beta\}, \beta>0$
  \item (saturated linear)
    $\xi\mapsto
    \left\{
    \begin{array}{ll}
    1, & \textrm{ if }\xi>1;\\
    \xi, & \textrm{ if }0\leq\xi\leq 1
  \end{array}\right.$
  \item (inverse square root unit)
    $\xi\mapsto \frac{\xi}{\sqrt{1+\xi^2}}$
  \item (arctangent)
    $\xi\mapsto (2/\pi)\arctan{\xi}$
  \item (hyperbolic tangent)
    $\xi\mapsto \tanh{\xi}$
  \item (inverse hyperbolic sine)
    $\xi\mapsto \arcsinh{\xi}$
  \item (Elliot)
    $\xi\mapsto \frac{\xi}{1+\xi}$
  \item (logarithmic)
    $\xi\mapsto \log(1+\xi)$
  \end{itemize}  
\item[(L2)] continuous scalar concave activation function satisfying $\lim_{\xi\to\infty}\sigma'(\xi)=1$:
  \begin{itemize}
  \item (ReLU, inverse square root linear unit)
    $\xi\mapsto \xi$    
  \end{itemize}
\end{enumerate}
\end{remark}

The above lists are by no means exhaustive. Indeed, a variety of the commonly used scalar activation functions reduce to a certain function from the above two lists if restricted to $\mathbb{R}_+$, so we do not list all variations here. For example, various versions of the ReLU (e.g., parameterized, exponential), including the original ReLU listed above, and also the inverse square root linear unit, reduce to the identity mapping on~$\mathbb{R}_+.$

The following fact enables us to establish scalability and strict monotonicity of activation functions, and it will be useful later in Section \ref{sect_construction}.
\begin{fact} \label{fact_act_fun}
Consider an $(A_{0,1,2})$-activation function $\sigma_i$ constructed with continuous scalar concave activation functions listed in Remark \ref{activ}. Then:
\begin{itemize}
\item if all scalar activation functions $\sigma_{i_j}$ are such that $\sigma_{i_j}(0)>0$ for $j\in\{1,2,\dots,k_i\}$, then, by \cite[Proposition~1]{cavalcante2015elementary}, $\sigma_i$ in (\ref{activ_layer}) is an $(A_{0,1,2,3})$-activation function. Thus, limiting the choice to scalar activation functions from lists (L1)-(L2) in Remark \ref{activ}, we obtain that $\sigma_i$ in (\ref{activ_layer}) is an $(A_{0,1,2,3})$-activation function only if it is constructed with the sigmoid scalar activation function; and
\item $\sigma_i$ in (\ref{activ_layer}) is strictly monotonic if it is constructed with scalar activation functions from lists (L1)-(L2) in Remark \ref{activ}, except for the capped ReLU and the saturated linear scalar activation functions.
\end{itemize}  
\end{fact}

\subsection{Construction of $(A_{0,1,2})$-neural networks} \label{sect_construction}
We now prove that the neural networks satisfying Assumption \ref{A012} are in the $(A_{0,1,2})$ class. Namely, the following lemma provides simple sufficient conditions to construct such networks. 


\begin{lemma} \label{construction}
Consider a neural network $T$ of the form in  (\ref{dnn}), and let $T_i\colon\sgens{k_{i-1}}\to\sgens{k_i}\colon\ x_{i-1}\mapsto\sigma_i(W_ix_{i-1}+b_i)$ be its $i$-th layer. Then:
\begin{enumerate}
\item if the affine mapping $y_i=W_ix_{i-1}+b_i$ has nonnegative weights $W_i$ and biases $b_i$, then it is in $(A_{0,1,2})$;
\item if the affine mapping $y_i=W_ix_{i-1}+b_i$ has nonnegative weights $W_i$ and positive biases $b_i$, then it is in $(A_{0,1,2,3})$;
\item if the affine mapping $y_i=W_ix_{i-1}+b_i$ is in $(A_{0,1,2})$ and the activation function $\sigma_i$ is also in $(A_{0,1,2})$, then the layer $T_i$ is in $(A_{0,1,2})$;
\item if the affine mapping $y_i=W_ix_{i-1}+b_i$ is in $(A_{0,1,2})$ and the activation function $\sigma_i$ is in $(A_{0,1,2,3})$, then the layer $T_i$ is in $(A_{0,1,2,3})$;
\item if the affine mapping $y_i=W_ix_{i-1}+b_i$ is in $(A_{0,1,2,3})$, and the activation function $\sigma_i$ is strictly monotonic and is in the class $(A_{0,2})$, then the layer $T_i$ is in $(A_{0,1,2,3})$;
\item if all layers $T_i$ for $i=1,2,\dots,n$ are in $(A_{0,1,2})$, then $T$ is an $(A_{0,1,2})$-neural network; 
\item if $T_{i_0}$ is in $(A_{0,1,2,3})$ for some $1\leq i_0\leq n$, $T_i$ are in $(A_{0,1,2})$ for $i=1,\dots,i_0-1$, and $T_i$ are strictly monotonic and are in $(A_{0,2})$ for $i=i_0+1,\dots,n$, then $T$ is an $(A_{0,1,2,3})$-neural network.
\end{enumerate}  
\end{lemma}
\proof See Appendix \ref{pd_construction}.

\begin{corollary} \label{ex1}
Suppose that a neural network $T$ with the general structure in (\ref{dnn}) satisfies Assumption \ref{A012}. Then:
\begin{enumerate}
\item $T$ is an $(A_{0,1,2})$-neural network; \label{ex1_1}
\item if, in addition, the $n$-th layer $T_n$ has a bias vector $b_n$ such that $b_n\gg 0$, and it uses a strictly monotonic activation function, then $T$ is an $(A_{0,1,2,3})$-neural network; or \label{ex1_2}
\item if, in addition, the $n$-th layer $T_n$ uses only $(A_{0,1,2,3})$ activation functions (e.g., sigmoid scalar activation functions), then $T$ is an $(A_{0,1,2,3})$-neural network. \label{ex1_3}
\end{enumerate}  
\end{corollary}
\proof By Assumption \ref{A012} and Lemma \ref{construction}, points \emph{1} and \emph{3}, the layers $T_i$ are in $(A_{0,1,2})$ for $i=1,\dots,n.$ By Lemma \ref{construction}, point \emph{6}, $T$ is an $(A_{0,1,2})$-neural network, which establishes point~\emph{1} above. Furthermore, if the additional assumption of point \emph{2} above is used, from Lemma \ref{construction}, points \emph{2} and \emph{5}, we obtain that the layer $T_n$ is in $(A_{0,1,2,3}).$ In Lemma \ref{construction}, point \emph{7}, we set $i_0=n$ to obtain the assertion of point \emph{2} above. Similarly, if the additional assumption of point \emph{3} above is used, from the first point in Fact~\ref{fact_act_fun} we obtain that $\sigma_n$ is an $(A_{0,1,2,3})$-activation function. Thus, from Lemma \ref{construction}, point \emph{4}, we conclude that $T_n$ is in $(A_{0,1,2,3}).$ For $i_0=n$, from Lemma \ref{construction}, point \emph{7}, we conclude that $T$ is an $(A_{0,1,2,3})$-neural network. $\blacksquare$

\subsection{Existence and uniqueness of fixed points of $(A_{0,1,2})$-neural networks} \label{sect.cond_fp}
Next, we prove the following key properties of $(A_{0,1,2})$-neural networks, which, as shown in the previous section, include as a subset neural networks satisfying Assumption \ref{A012}:
\begin{enumerate}
\item We provide conditions for the existence of fixed points of an $(A_{0,1,2})$-neural network
(Corollary \ref{nonemptylo}).
\item We provide conditions for the uniqueness of the fixed point of a neural network~$T$ in cases where $T$ is:
\begin{enumerate}
\item an $(A_{0,1,2,3})$-neural network (Corollary \ref{uniqulo}),
\item a strongly primitive\footnote{The definition of strong primitivity is given in Definition \ref{primitivo} below.} $(A_{0,1,2})$-neural network, and with its activation functions constructed with the continuous scalar concave activation functions listed in Remark \ref{activ} (Corollary~\ref{cor.uniqueness}). 
\end{enumerate}
\item We show that the condition for the existence of a fixed point of an $(A_{0,1,2})$-neural network (Corollary \ref{nonemptylo}) is weaker than those obtained using arguments in convex analysis (Remark \ref{weak_weights}).
\item We provide numerically efficient methods for verifying the existence and uniqueness of fixed point(s) of $(A_{0,1,2})$-neural networks (Proposition~\ref{proposition.primitive} and Corollary \ref{cor.uniqueness}).   
\end{enumerate}

\subsubsection{Nonlinear spectral radius of an $(A_{0,1,2})$-neural network} \label{radius}
The path toward obtaining the main results of Section \ref{sect.cond_fp} is via the concept of the asymptotic mapping associated with an $(A_{0,1,2})$-neural network (Definition~\ref{asym_map}), and we provide a closed algebraic form of the asymptotic mapping of an $(A_{0,1,2})$-neural network in Proposition \ref{sr}. This result is then used to determine the (nonlinear) spectral radius of an $(A_{0,1,2})$-neural network in Corollary \ref{sr_cor}.

\begin{definition} \label{asym_map}
Let $T:\sgens{k}_+\to\sgens{k}_+$ be an $(A_{0,1,2})$-neural network of the form (\ref{dnn}). The asymptotic mapping associated with $T$ is the mapping defined by
\begin{equation}
T_\infty:\sgens{k}_+\to\sgens{k}_+:x\mapsto \lim_{p\to\infty}\dfrac{1}{p}T(p x).
\end{equation}
We recall that the above limit always exists \cite{Cavalcante2019_IEEE_TSP}.
\end{definition}


We now prove that the asymptotic mapping $T_\infty$ associated with an $(A_{0,1,2})$-neural network $T$ is linear, even if the activation functions are possibly nonlinear. This property is very special because asymptotic mappings are nonlinear in general \cite{Cavalcante2019_IEEE_TSP}, and linearity is a property that can be exploited to simplify the analysis of the set of fixed points of $(A_{0,1,2})$-neural networks. 



For the common case of a nonnegative neural network with activation functions constructed with continuous scalar concave activation functions listed in Remark~\ref{activ}, we have the following result:

\begin{proposition} \label{sr}
Consider a neural network $T$ of the form in (\ref{dnn}) with nonnegative weight matrices and biases. Assume that the activation functions are constructed with continuous scalar concave activation functions from lists (L1)-(L2) in Remark \ref{activ}. Then $T$ is an $(A_{0,1,2})$-neural network. Moreover:
\begin{enumerate}
\item If all layers use continuous scalar concave activation functions from list (L2) in Remark~\ref{activ}, then the asymptotic mapping of $T$ is given by
  $T_\infty(x)=\prod_{i=n}^1W_ix.$ \label{sr_1}
\item On the other hand, if at least one layer of $T$ uses only continuous scalar concave activation functions from list~(L1) in Remark \ref{activ}, then $T_\infty(x)=0.$ \label{sr_2}
\end{enumerate}  
\end{proposition}  
\proof From Remark \ref{activ} we obtain that $\sigma_i$ are $(A_{0,1,2})$-activation functions for $i=1,2,\dots,n.$ Thus, $T$ satisfies both parts of Assumption \ref{A012}, and, hence, from Corollary \ref{ex1}, point \emph{1}, we obtain that $T$ is an $(A_{0,1,2})$-neural network.

\emph{Case 1) :} Define
\begin{equation*}
T_{1_\infty}(x)\df\lim_{p\to\infty}\frac{1}{p}T_1(px)=\lim_{p\to\infty}\frac{\sigma_1(W_1px+b_1)}{p}.
\end{equation*}
Applying L'H\^{o}pital's rule to the coordinate continuous scalar concave activation functions $(\sigma_{1_j}:\mathbb{R}\to \mathbb{R})_{j\in\{1,\ldots,k_1\}}$ from list (L2) in Remark \ref{activ}, with the scalar $p$ as their arguments with $x$ fixed, we deduce
\begin{equation*}
T_{1_\infty}(x)=\lim_{p\to\infty}\sigma'_1(W_1px+b_1)W_1x=W_1x.
\end{equation*}
Analogously, for the second layer we have
\begin{equation*}
T_{{1,2}_\infty}(x)\df\lim_{p\to\infty}\frac{1}{p}T_2(T_1(px))=\\\lim_{p\to\infty}\frac{\sigma_2\Big(W_2\big(\sigma_1(W_1px+b_1)\big)+b_2\Big)}{p}.
\end{equation*}
Then, applying L'H\^{o}pital's rule to the coordinate functions $(\sigma_{2_j}:\mathbb{R}\to \mathbb{R})_{j\in\{1,\ldots,k_2\}}$ from list (L2) in Remark \ref{activ}, we obtain 
\begin{equation} \label{two_layers_as}
T_{{1,2}_\infty}(x)=\lim_{p\to\infty}\sigma'_2\Big(W_2\big(\sigma_1(W_1px+b_1)\big)+b_2\Big)W_2\sigma'_1(W_1px+b_1)W_1x=W_2W_1x.
\end{equation}
Proceeding as above for the remaining layers, we conclude that $T_\infty(x)=\prod_{i=n}^1W_ix.$

\emph{Case 2) :} If $\sigma_i$ is composed only of continuous scalar concave activation functions from list (L1) in Remark \ref{activ} for some $1\leq i\leq n$, then $\lim_{p\to\infty}\sigma'_i(p)=0.$ In such a case, we obtain $T_\infty(x)=0$, as can be deduced from the special case $n=2$ in (\ref{two_layers_as}). $\blacksquare$

The asymptotic mapping obtained in Proposition \ref{sr} associated with an $(A_{0,1,2})$-neural network $T$ can be used to define the (nonlinear) spectral radius of $T$ as follows:

\begin{definition} \label{def.spec_rad}
The (nonlinear) spectral radius of an $(A_{0,1,2})$-neural network $T$ of the form (\ref{dnn}) is given by the largest eigenvalue of the corresponding (linear) asymptotic mapping \cite{Oshime1992,Cavalcante2019_IEEE_TSP}:
  \begin{equation*}
    \rho(T_\infty)=\max\{\lambda\in\mathbb{R}_+:\ \exists\ x\in\sgens{k}_+\backslash\{0\}\textrm{ s.t. }T_\infty(x)=\lambda x\}\in\mathbb{R}_+.
  \end{equation*}  
\end{definition}

From Proposition \ref{sr} and Definition \ref{def.spec_rad} we obtain the following corollary yielding the desired result of this section.

\begin{corollary} \label{sr_cor}
Consider a neural network $T$ of the form (\ref{dnn}) with nonnegative weight matrices and biases. Assume that the activation functions are constructed with continuous scalar concave activation functions from lists (L1)-(L2) in Remark \ref{activ} [from Proposition \ref{sr}, $T$ is an $(A_{0,1,2})$-neural network]. Then the spectral radius of $T$ is given by   
\begin{equation} \label{fix2}
\rho(T_\infty)=\max\{\lambda\in\mathbb{R}_+:\  \exists\ x\in\sgens{k}_+\backslash\{0\}\textrm{ s.t. } \prod_{i=n}^1W_ix=\lambda x\}
\end{equation}
if all layers $i=1,2,\dots,n$ use continuous scalar concave activation function from list~(L2) in Remark \ref{activ}. On the other hand, if at least one layer of~$T$ uses only continuous scalar concave activation functions from list (L1) in Remark~\ref{activ}, then $\rho(T_\infty)=0.$ 
\end{corollary}

\subsubsection{Conditions for existence of fixed points of $(A_{0,1,2})$-neural networks} \label{existence}
The following corollary of Proposition \ref{P1_R2019} gives a sufficient condition for $(A_{0,1,2})$-neural networks to have a nonempty fixed point set.

\begin{corollary} \label{nonemptylo}
Consider an $(A_{0,1,2})$-neural network $T$ of the form (\ref{dnn}). Assume that $\rho(T_\infty)<1.$ Then, from Proposition \ref{P1_R2019}, we have that $\Fix{T}\neq\emptyset.$ 
\end{corollary}

Furthermore, the following corollary shows that for $(A_{0,1,2,3})$-neural networks, the assertion of Corollary \ref{nonemptylo} can be significantly strengthened. More precisely, in this setting the condition $\rho(T_\infty)<1$  is not only sufficient but also necessary to prove the existence and uniqueness of the fixed point. 

\begin{corollary} \label{uniqulo}
Consider an $(A_{0,1,2,3})$-neural network $T$ of the form (\ref{dnn}). Then, from Proposition \ref{P4_R2019_TSP}, the neural network $T$ has a fixed point if and only if $\rho(T_\infty)<1.$ Furthermore, from Proposition \ref{F1_R2019_TSP} the fixed point is unique, and it is also positive as a consequence of the scalability of $T.$
\end{corollary}

\subsubsection{Comparison of the assumptions implying the existence of fixed points of $(A_{0,1,2})$-neural networks} \label{comparison}
In this subsection, we emphasize that the results of Proposition~\ref{sr}, Corollary \ref{nonemptylo}, and Corollary \ref{uniqulo} yield weaker conditions for the existence of fixed points of $(A_{0,1,2})$-neural networks than existing conditions based on arguments in convex analysis. To establish this claim, we use the following fact.

\begin{fact} \label{nonexp}
Consider a neural network $T$ of the form (\ref{dnn}) with nonnegative weight matrices and biases. Assume that the activation functions are constructed with continuous scalar concave activation functions from lists (L1)-(L2) in Remark \ref{activ} [from Proposition \ref{sr}, $T$ is an $(A_{0,1,2})$-neural network]. A standard result is that the spectral radius of a matrix, in the conventional sense in linear algebra, is the infimum of all matrix operator norms,  so the spectral radius $\rho(W)$ of the nonnegative matrix $W$, given by
\begin{equation*}
\sgen{k}{k}\ni W\df \prod_{i=n}^1W_i,
\end{equation*}
can be equivalently written as $\rho(W)=\inf\{\|W\|:\ \|\cdot\|\textrm{ - operator norm of }W\}$.  Therefore, if all layers $i=1,2,\dots,n$ use continuous scalar concave activation function from list (L2) in Remark \ref{activ}, from Corollary \ref{sr_cor} and arguments in classical Perron-Frobenius theory \cite[p. 670]{meyer2000matrix} we obtain that
\begin{equation} \label{fixxx}
\rho(T_\infty)=\rho(W)\leq\|W\|\leq\prod_{i=1}^n\|W_i\|,
\end{equation}  
where the matrix norm $\|\cdot\|$ in (\ref{fixxx}) is an arbitrary operator norm. The second inequality follows from the submultiplicativity property of operator norms.
\end{fact}

We are now able to provide concrete examples of $(A_{0,1,2})$-neural networks for which the existence of fixed points can be established using the above results, but not from previous arguments based on convex analysis. We summarize them in the following remark.

\begin{remark} \label{weak_weights}
Let $T$ be a neural network of the form in (\ref{dnn}) with nonnegative weight matrices and biases. Assume that the activation functions are selected from the continuous scalar concave activation functions in lists (L1)-(L2) in Remark \ref{activ} [from Proposition \ref{sr}, $T$ is an $(A_{0,1,2})$-neural network]. Then:
\begin{itemize}
\item From Proposition \ref{sr} and Corollary \ref{nonemptylo}, we conclude that if at least one layer of $T$ uses only continuous scalar concave activation functions from list (L1) in Remark~\ref{activ}, then $T$ has a nonempty fixed point set for arbitrary (finite) nonnegative weights. No assumptions on the (operator or other) norms of the weight matrices are necessary to guarantee the existence of fixed points.
\item From Fact \ref{nonexp}, we obtain that, even if all layers $i=1,2,\dots,n$ use exclusively continuous scalar concave activation functions from list (L2) in Remark \ref{activ}, the conditions for the existence of a nonempty fixed point set stated in Corollaries~\ref{nonemptylo} and \ref{uniqulo} are weaker than those obtained using arguments in convex analysis \cite{Combettes2020a} [see also the bounds on the Lipschitz constant of a neural network with nonnegative weights in \cite[Section 5.3]{Combettes2020b}]. 
\end{itemize}  
\end{remark}

\subsubsection{Numerically efficient methods for verifying the existence of fixed points of strongly primitive $(A_{0,1,2})$-neural networks}
We now introduce the concepts of primitivity and strong primitivity of $(A_{0,1,2})$-neural networks. The former is used in Section \ref{appnn},\footnote{We present both definitions here because they are closely related.} and the latter is used in Proposition~\ref{proposition.primitive} and Corollary \ref{cor.uniqueness} below to obtain numerically efficient methods for verifying the existence and uniqueness of fixed points of neural networks. 

\begin{definition} \label{primitivo}
We say that an $(A_{0,1,2})$-neural network $T\colon\sgens{k}_+\to \sgens{k}_+$ is primitive if
\begin{equation*}
\forall x\in\sgens{k}_+\backslash\{0\} \quad \exists m\in\mathbb{N}\textrm{ s.t. }T^m(x)\gg 0.
\end{equation*}
Similarly, we say that $T$ is strongly primitive if
\begin{equation*}
\forall x\in\sgens{k}_+\quad \exists m\in~\mathbb{N}\textrm{ s.t. }T^m(x)\gg 0.
\end{equation*}  
\end{definition}

\begin{proposition} \label{proposition.primitive}
Consider an $(A_{0,1,2})$-neural network $T\colon\sgens{k}_+\to \sgens{k}_+.$ In addition, assume that there exists $m\in\mathbb{N}$ such that $T^m(0) \gg 0.$ Then each of the following holds:
\begin{enumerate} 
       \item $T$ is strongly primitive.
       \item If $\rho(T_\infty)<1$, then $\emptyset\neq\mathrm{Fix}(T)\subset \op{int}(\sgens{k}_+)$. In contrast, if $\rho(T_\infty)>1$, then $\mathrm{Fix}(T)=\emptyset$.
       \item If $H:\sgens{k}_+\to \sgens{k}_+:x\mapsto T^m(x)$ is an $(A_{0,1,2,3})$-neural network and $T$ is as in Proposition~\ref{sr}, then $T$ has a fixed point if and only if $\rho(T_\infty)<1.$ Furthermore, $\mathrm{Fix}(T)=\mathrm{Fix}(H)$, and the fixed point of $T$ is unique and positive.
       \item If $H:\sgens{k}_+\to \sgens{k}_+:x\mapsto T^m(x)$ is an $(A_{0,1,2,3})$-neural network and $T$ is as in Proposition~\ref{sr} with $\rho(T_\infty)<1$, then the sequence $(x_n)_{n\in\mathbb{N}}$ generated via $x_{n+1}=T(x_n)$, with $x_1\in\sgens{k}_+$ arbitrarily chosen, converges to the unique fixed point of $T$.
\end{enumerate}
\end{proposition}
\proof 1) By assumption, $\exists m\in\mathbb{N}$ such that $T^m(0) \gg 0.$ Let $x\in\sgens{k}_+.$ From the monotonicity of $T$, one has $\forall n\ge m\ T^n(x)\ge T^n(0)\gg 0$, and the desired result follows.

2) The first statement is immediate from Corollary~\ref{nonemptylo} and strong primitivity of $T.$ Now, for the sake of contradiction, assume that $\rho(T_\infty)>1$ and the network has a fixed point denoted by $x\in\mathrm{Fix}(T)$. Part 1) of the proposition shows that $x=T^m(x)\gg 0$, which is a contradiction because \cite[Proposition~2]{Cavalcante2019} asserts that a monotonic and weakly scalable $T$ with $\rho(T_\infty)>1$ has no fixed points in $\op{int}(\sgens{k}_+).$

3) From the definition of $H$, we verify that $\mathrm{Fix}(T)\subseteq \mathrm{Fix}({H})$, because if $x^\star$ is such that $T(x^\star)=x^\star$, then $H(x^\star)=T^m(x^\star)=x^\star.$ By repeating the proof of Proposition \ref{sr} for $H$, we obtain that $H_\infty(x)=W^m(x)$ if and only if $T_\infty(x)=\prod_{i=n}^1W_ix$, where $W\df \prod_{i=n}^1W_i$, and $H_\infty(x)=0$ if and only if $T_\infty(x)=0.$ Then, for the first case, the equality $\rho(H_\infty)=(\rho(T_\infty))^m$ is obtained from \cite[Theorem~1]{Johnson1990}, whereas it is immediate for the second case. Hence, in particular, $\rho(T_\infty)<1$ if and only if $\rho(H_\infty)<1.$

Now consider the case $\rho(T_\infty)<1$. Then, we conclude that the extended neural network $H$ has a unique (positive) fixed point as a consequence of the statement in Corollary~\ref{uniqulo}. Moreover, Part 2) of the proposition guarantees that $\mathrm{Fix}(T)\neq\emptyset$, so we must have $\mathrm{Fix}(H)=\mathrm{Fix}(T)$ because $\mathrm{Fix}(H)$ is a singleton and $\emptyset\neq \mathrm{Fix}(T)\subseteq\mathrm{Fix}(H)$. Similarly, for the case $\rho(T_\infty)\ge 1$, we deduce $\rho(H_\infty)=(\rho(T_\infty))^m\ge 1$, which implies $\mathrm{Fix}(H)=\emptyset$ from the statement of Corollary~\ref{uniqulo}. Therefore, $T$ cannot have a fixed point because $\mathrm{Fix}(T)\subseteq \mathrm{Fix}({H})=\emptyset$, and the proof is complete. 

4) Part 3) of the proposition and the assumptions of Part 4) guarantee that $T$ and $H$ have a common unique positive fixed point that we denote by $x^\star\in\op{int}(\sgens{k}_+)$. We also note that, as $T$ is monotonic and weakly scalable on $\op{int}(\sgens{k}_+)$, by \cite[Lemma 2.1.7]{Lemmens2012} it is nonexpansive in the Thompson metric space $(\Intr, d_T)$, see Definition~\ref{tmetric}.
Let $i\in\mathbb{N}$ and let $k\geq im.$ Then
\begin{equation*}
\forall x\in\sgens{k}_+\quad\forall k\geq im\quad d_T(T^k(x),x^\star)=d_T(T^k(x),T^{k-im}(x^\star))\leq d_T(T^{im}(x),x^\star),
\end{equation*}  
where the last inequality follows from the nonexpansivity of $T$ w.r.t. $d_T.$ We note that for $k\geq im$ if $i\to\infty$ then $k\to\infty.$ Thus, we obtain that
\begin{equation*}
\forall x\in\sgens{k}_+\quad \lim_{k\to\infty}d_T(T^k(x),x^\star)\leq\lim_{i\to\infty}d_T(T^{im}(x),x^\star)=\lim_{i\to\infty}d_T(H^{i}(x),x^\star)=0,
\end{equation*}
where the last equality follows from \cite[Fact 1(iv)]{Cavalcante2019_IEEE_TSP}. $\blacksquare$

One of the practical implications of Proposition~\ref{proposition.primitive} is that we can obtain a simple numerical certificate of uniqueness of the fixed point of a monotonic and weakly scalable nonnegative neural network $T$ with concave activation functions, by simply computing $T^m(0)$ for some sufficiently large $m$, as formally described below. It is also important to note that in such a case, the network $T$ needs only to be monotonic and weakly scalable to have a unique fixed point if $\rho( T_\infty)<1$, which is a stronger result than the one given in Corollary~\ref{uniqulo}, which requires scalability of the network.

\begin{corollary} \label{cor.uniqueness}
Consider a neural network $T$ of the form (\ref{dnn}) with nonnegative weight matrices and biases. Assume that the activation functions are selected from the continuous scalar concave activation functions in lists (L1)-(L2) in Remark \ref{activ} [from Proposition \ref{sr}, $T$ is an $(A_{0,1,2})$-neural network]. If there exists $m\in\mathbb{N}$ such that $T^m(0)\gg 0$, then $T$ has a fixed point if and only if $\rho(T_\infty)<1.$ Furthermore, the fixed point is unique and positive.
\end{corollary} 
\proof Let $f_1,f_2\colon\sgens{k_1}_+\to\sgens{}_+$ be nonnegative concave functions and let $t\in (0,1).$ We note that, by [Lemma 1]\cite{Cavalcante2019}, they are $(A_{0,1,2})$-functions. From concavity of $f_1$ one has
\begin{equation*}
f_1(tx_1 + (1-t)x_2) \geq tf_1(x_1) + (1-t)f_1(x_2).
\end{equation*}
Hence, monotonicity of $f_2$ implies that
\begin{equation*}
f_2[f_1(tx_1 + (1-t)x_2)] \geq f_2[tf_1(x_1) + (1-t)f_1(x_2)].
\end{equation*}
From the concavity of $f_2$ we deduce
\begin{equation*}
f_2[tf_1(x_1) + (1-t)f_1(x_2)]\geq tf_2(f_1(x_1))+(1-t)f_2(f_1(x_2)).
\end{equation*}
By combining the above two inequalities, we conclude that
\begin{equation*}
(f_2\circ f_1)(tx_1 + (1-t)x_2)\geq t(f_2\circ f_1)(x_1)+(1-t)(f_2\circ f_1)(x_2),
\end{equation*}
establishing that $f_2\circ f_1$ is also a nonnegative concave function.  We can use induction to extend the validity of the above arguments to an arbitrary number of nonnegative concave functions. By definition, a mapping $F\df(f_1\dots,f_{k_2})\colon\sgens{k_1}_+\to\sgens{k_2}_+$ is said to be nonnegative concave if all coordinate functions $(f_1\dots,f_{k_2})$ are nonnegative concave functions. Thus, we conclude that the layers of $T$ are nonnegative concave mappings because they are constructed by composing nonnegative affine mappings with nonnegative concave activation functions. Hence, the neural network $T$ is itself a nonnegative concave mapping as a composition of $n$ nonnegative concave layers. Furthermore, $H:\sgens{k}_+\to\sgens{k}_+: x\mapsto T^m(x)$ is a nonnegative concave mapping because it is the $m$-fold composition of $T$ with itself. We also deduce that $H$ is necessarily a positive mapping because, by the monotonicity of $T$, $\forall x\in\sgens{k}_+~H(x)= T^m(x)\ge T^m(0)\gg 0$. Concavity and positivity of $H$ in the nonnegative orthant imply that $H$ is an $(A_{0,1,2,3})$-neural network \cite[Proposition~1]{cavalcante2015elementary}. Thus, from Proposition~\ref{proposition.primitive}, point 3, we obtain that $T$ has a fixed point if and only if $\rho(T_\infty)<1$. Furthermore, this fixed point is unique and positive. $\blacksquare$ 

\subsection{The shape of fixed point sets of $(A_{0,1,2})$-neural networks}
In the previous sections, we provided conditions for the existence and uniqueness of fixed points of $(A_{0,1,2})$-neural networks under different mild assumptions. However, if the fixed point set is not a singleton, we have not yet established the shape of this set. In the following, we address this limitation of the analysis done so far. In more detail, in this section, we show the following results:
\begin{enumerate}
\item First, in Section \ref{nPFT}:
\begin{enumerate}
\item in Proposition \ref{p3}, we show the shape of the fixed point set of a generic $(A_{0,1,2})$-mapping under the assumption of lower- and upper-primitivity (see Definition \ref{L_U_defs} below) at its fixed points; and
\item we show that the condition of lower- and upper-primitivity is essential to obtain a meaningful shape of the fixed point set of a generic $(A_{0,1,2})$-mapping, otherwise the fixed point set of such a mapping can be in principle arbitrary (Examples \ref{ex_L_U_1} and \ref{ex_L_U_2}).
\end{enumerate}
\item Afterwards, in Section \ref{appnn}, we specialize the results of Section \ref{nPFT} to the case of $(A_{0,1,2})$-neural networks. More precisely:
\begin{enumerate}
\item we compare the results in Section \ref{existence} and Section \ref{nPFT} related to the shape of the fixed point set of $(A_{0,1,2})$-neural networks (Remark \ref{yar});
\item we derive a numerically verifiable sufficient condition for an $(A_{0,1,2})$-neural network to be lower- and upper-primitive at its fixed points (Lemma \ref{suff_L_U}).
\end{enumerate}  
\end{enumerate}

\subsubsection{Nonlinear Perron-Frobenius Theory} \label{nPFT}
The results of this section establish the shape of the fixed point sets of $(A_{0,1,2})$-mappings $f\colon\sgens{k}_{+}\to\sgens{k}_{+}$ in the interior of the $\sgens{k}_{+}$ cone. They are derived using tools from nonlinear Perron-Frobenius theory, and the following definition is required to state the main contributions:

\begin{definition}\cite[Definition 6.5.2]{Lemmens2012} \label{L_U_defs}
Let $f:\op{int}(\sgens{k}_+)\to\op{int}(\sgens{k}_+).$ We say that $f$ is lower-primitive at $x\in\op{int}(\sgens{k}_+)$ if there exists an open neighbourhood $U\subseteq\op{int}(\sgens{k}_+)$ of $x$ such that for each $y\in U$ with $y<x$ there exists an integer $p_y\geq 1$ with $f^{p_y}(y)\ll f^{p_y}(x).$ Likewise, we say that $f$ is upper-primitive at $x\in\op{int}(\sgens{k}_+)$ if there exists an open neighbourhood $U\subseteq\op{int}(\sgens{k}_+)$ of $x$ such that for each $y\in U$ with $x<y$ there exists an integer $p_y\geq 1$ with $f^{p_y}(x)\ll f^{p_y}(y).$    
\end{definition}

The following proposition provides the shape of the fixed point set of an $(A_{0,1,2})$-mapping under the conditions of lower- and upper-primitivity. It
extends \cite[Proposition 3]{Cavalcante2019} to the case $x\in\op{int}(\sgens{k}_+)$, and it can be seen as a corollary of \cite[Theorem 6.5.6]{Lemmens2012}. For clarity, we provide a proof that quotes parts of the original proof of \cite[Theorem 6.5.6]{Lemmens2012}.

\begin{proposition}[\emph{cf.} Theorem 6.5.6 in \cite{Lemmens2012}] \label{p3}
Let us define $f:\op{int}(\sgens{k}_+)\to\op{int}(\sgens{k}_+)$ to be the restriction of an $(A_{0,1,2})$-mapping to the positive orthant. Furthermore, let $\Fix{f}\neq\emptyset$ and let $f$ satisfy the conditions of lower- and upper-primitivity at $u\in\Fix{f}.$ Let $x\in\op{int}(\sgens{k}_+).$ Then, there exists $\lambda_x>0$ such that $\lim_{p\to\infty}f^p(x)=\lambda_xu\in\Fix{f}.$ Moreover, define
\begin{equation*}
s_0\df\inf\{\lambda>0:\ \lambda u\in\Fix{f}\},
\end{equation*}  
and
\begin{equation*}
t_0\df\sup\{\lambda>0:\ \lambda u\in\Fix{f}\}.
\end{equation*}  
Then:
\begin{align} \label{fixtint}
   \begin{split}
     & \textrm{if } s_0>0 \textrm{ and } t_0<\infty\Rightarrow\ \Fix{f}=\{\lambda u:\ s_0\leq\lambda\leq t_0\};\\
     & \textrm{if } s_0=0 \textrm{ and } t_0<\infty\Rightarrow\ \Fix{f}=\{\lambda u:\ 0<\lambda\leq t_0\};\\
     & \textrm{if } s_0>0 \textrm{ and } t_0=\infty\Rightarrow\ \Fix{f}=\{\lambda u:\ s_0\leq\lambda\}; \text{ and} \\
     & \textrm{if } s_0=0 \textrm{ and } t_0=\infty\Rightarrow\ \Fix{f}=\{\lambda u:\ 0<\lambda\}.\\
   \end{split}
\end{align}
\end{proposition}
\proof Let $u\in\Fix{f}$ and let $x\in\op{int}(\sgens{k}_+).$ As $f$ is monotonic and weakly scalable on $\op{int}(\sgens{k}_+)$, by \cite[Lemma 2.1.7]{Lemmens2012} it is nonexpansive in the Thompson metric space $(\Intr, d_T)$ (see Definition~\ref{tmetric}), which is a complete metric space \cite[Proposition 2.5.2]{Lemmens2012}. We further note from \cite[Remark 3 and Corollary 1]{Piotrowski2022} that $(\Intr, d_T)$ is also a proper metric space in the sense that every closed ball in $(\Intr, d_T)$ is compact.

Denote the orbit of $f$ at $x$ by $\cal{O}(x;f)\df\{f^p(x):\ p=0,1,2,\dots\}$ with $f^0(x)\df x.$ Then, the orbit $\cal{O}(u;f)=\{u\}$ is bounded, thus, by Całka's Theorem \cite[Theorem 3.1.7]{Lemmens2012}, $\cal{O}(x;f)$ is also bounded in $(\Intr, d_T).$ The closure $\closure{\cal{O}}(x;f)$ of $\cal{O}(x;f)$ consists of $\cal{O}(x;f)$ and the bounded set of all limit points of subsequences of $\cal{O}(x;f).$ Hence, by \cite[Corollary 1]{Piotrowski2022}, $\closure{\cal{O}}(x;f)$ is compact in $(\Intr, d_T).$ Then, from \cite[Lemma 3.1.2]{Lemmens2012}, one has that $f(\omega(x;f))=\omega(x;f)$, where $\omega(x;f)\df\{y\in\op{int}(\sgens{k}_+):\ f^{p_i}(x)\to y\textrm{ for some }(p_i)_i\textrm{ with }p_i\to\infty\}$, and, consequently, $f^p(\omega(x;f))=\omega(x;f)$ for any $p\in\mathbb{N}.$

Furthermore, by our assumptions on $f$, we can apply \cite[Lemma~6.5.4]{Lemmens2012} to $f$ and conclude that there are two cases: either there exists $0<\alpha_x\leq 1$ such that $m(y/u)=\alpha_x$ for all $y\in\omega(x;f)$, or there exists $\beta_x\geq 1$ such that $M(y/u)=\beta_x$ for all $y\in\omega(x;f)$, where $m(y/u)\df\sup\{\gamma>0:\ \gamma u\leq y\}$ and $M(y/u)\df\inf\{\gamma>0:\ y\leq \gamma u\}.$ We note that for $y,u\in\op{int}(\sgens{k}_+)$ one has simply $m(y/u)=\min_{i=1,\dots,k}\{y_i/u_i\}$ and $M(y/u)=\max_{i=1,\dots,k}\{y_i/u_i\}$, see \cite[eqs.(2.8),(2.9)]{Lemmens2012}.

We now show that, in the first case, $\omega(x;f)=\{\alpha_x u\}.$ To obtain a contradiction, we assume that there exists $z\in\omega(x;f)$ with $z\neq \alpha_x u$ such that $m(z/u)=\alpha_x.$
Then $z_i\geq\alpha_x u_i$ for $i=1,\dots,k$, and there exists $i_0\in\{1,\dots,k\}$ such that $z_{i_0}>\alpha_x u_{i_0}$, hence $\alpha_x u<z.$ Since $f$ is upper-primitive at $u$, there exists $p\in\mathbb{N}$ such that
\begin{equation*}
\alpha_x u=\alpha_x f^p(u)\ll\alpha_x f^p(\alpha_x^{-1}z)\leq f^p(z),
\end{equation*}  
where the second inequality follows from the weak scalability of $f.$ Consequently, $\alpha_x u\ll f^p(z).$ It follows that $m(f^p(z)/u)=\sup\{\gamma>0:\ \gamma u\leq f^p(z)\}>\alpha_x$, which is impossible, as $f^p(z)\in\omega(x;f).$ In a similar way, it can be shown that, in the second case, $\omega(x;f)=\{\beta_x u\}.$ Thus, we have proved that there exists $\lambda_x>0$ such that $\omega(x;f)=\{\lambda_x u\}.$

Consequently, since $f(\omega(x;f))=\omega(x;f)$, we have $f(\lambda_x u)=\lambda_x u$, thus $\lambda_x u\in\Fix{f}.$ We also note that, if $v\in\Fix{f}$ is such that $v\neq u$, we obtain that $\omega(v;f)=\{v\}$, and thus it must be either $v=\alpha_v u$ or $v=\beta_v u$, where $0<\alpha_v\leq 1$ and $1\leq\beta_v.$ 

Furthermore, as $\alpha_x u\leq u\leq\beta_x u$ for each $x\in\op{int}(\sgens{k}_+)$, from \cite[Lemma~6.5.5]{Lemmens2012} we obtain that $\Fix{f}$ is in this case one of the intervals in (\ref{fixtint}). Moreover, from \cite[Lemma~3.1.3]{Lemmens2012} for a periodic point with period~1, one has that
\begin{equation*}
\lim_{p\to\infty}f^p(x)=\lambda_x u\in\Fix{f}.\ \blacksquare 
\end{equation*}

The conditions of lower- and upper-primitivity are important, as the following examples demonstrate.

\begin{example} \label{ex_L_U_1}
Let $f:\op{int}(\sgens{2}_+)\to\op{int}(\sgens{2}_+)$ be defined as $f(x)=(x_1,\min\{2,x_2\})$ for $x=(x_1,x_2)\in\op{int}(\sgens{2}_+).$ Then $f$ is an $(A_{0,1,2})$-mapping that is also subadditive, see \cite[Example 2.4]{Oshime1992}. We can check that $\Fix{f}=\op{int}(\mathbb{R}_+)\times (0,2].$ However, neither of the conditions of lower- and upper-primitivity are satisfied by $f$ on $\Fix{f}$, as the following analysis shows.

Let $u=(u_1,u_2)\in\Fix{f}$, and let $x=(2u_1,u_2).$ Then $x\in\Fix{f}$ and $u<x$ because $x\neq u$ and $x-u=(u_1,0)\in\sgens{2}_{+}.$ However, for all $p_x\geq 1$, we have $f^{p_x}(u)=u\centernot{\ll}f^{k_x}(x)=x$ because $x-u=(u_1,0)\notin\op{int}(\sgens{2}_+).$ Similarly, if $x=(u_1/2,u_2)\in\Fix{f}$, then $x<u$ because $u\neq x$ and $u-x=(u_1/2,0)\in\sgens{2}_{+}$. However, for all $p_x\geq 1$, we have $f^{p_x}(x)=x\centernot{\ll}f^{p_x}(u)=u$ because $u-x=(u_1/2,0)\notin\op{int}(\sgens{2}_+).$
\end{example}

\begin{example} \label{ex_L_U_2}
Let $f:\op{int}(\sgens{2}_+)\to\op{int}(\sgens{2}_+)$ be the Gauss arithmetic-geometric mean given by $f(x)=(\frac{x_1+x_2}{2},\sqrt{x_1x_2})$ for $x=(x_1,x_2)\in\op{int}(\sgens{2}_+).$ Then $f$ is an $(A_{0,1,2})$-mapping, and, for each $x\in\op{int}(\sgens{2})$, there exists $\lambda>0$ such that $\lim_{p\to\infty}f^p(x_1,x_2)=(\lambda,\lambda)$ \cite[Section 1.4]{Lemmens2012}. Indeed, it is clear that $\Fix{f}=\{x\in\op{int}(\sgens{2}_+)\ |\ x_1=x_2\}$, and, from the properties of arithmetic and geometric means, we verify that $f$ is both lower- and upper-primitive at each $u\in\Fix{f}.$ Hence, let $u=(\lambda,\lambda)$ for some $\lambda>0$, and let $x\in\op{int}(\sgens{2}_+).$ Then, from Proposition \ref{p3} we conclude that there exists $\lambda_x>0$ such that $\lim_{p\to\infty}f^p(x)=(\lambda_x\lambda,\lambda_x\lambda).$ It is also seen that, in this case, $\Fix{f}=\{(\gamma,\gamma),\ \gamma>0\}$, see~(\ref{fixtint}).
\end{example}

We close this section with the following remarks, which will be useful later in this study.

\begin{remark} \label{lu_primitive}
From the proof of Proposition \ref{p3} it is seen that, if $f$ is both lower- and upper-primitive at any $u\in\Fix{f}$, then it satisfies these conditions on the whole $\Fix{f}.$
\end{remark}

\begin{remark} \label{extension}
Every $(A_{0,1,2})$-mapping $f:\op{int}(\sgens{k}_+)\to\op{int}(\sgens{k}_+)$ has a continuous extension $f_{\op{ext}}\colon\sgens{k}_{+}\to\sgens{k}_{+}$ that is also an $(A_{0,1,2})$-mapping \cite[Theorem 3.10]{Burbanks2003}. This fact justifies calling $f$ itself an $(A_{0,1,2})$-mapping in the sense of Definition \ref{main_df}.
\end{remark}

\subsubsection{Applications to neural networks} \label{appnn}
We now specialize the results derived above to neural networks.

\begin{remark} \label{yar}
From Corollary \ref{uniqulo}, we verify that, if a neural network $T$ of the form (\ref{dnn}) is in $(A_{0,1,2,3})$, then $\Fix{T}$ is either the empty set or a singleton. Moreover, from Corollary~\ref{nonemptylo} and Proposition~\ref{p3}, we conclude that $\Fix{T}$ remains simple if $T$ is more generally an $(A_{0,1,2})$-neural network, because in this case $\Fix{T}$ is an interval. This result is obtained in Proposition \ref{p3} under the assumption that $T$ grows ``fast enough'' at its fixed points, i.e., it satisfies the conditions of lower- and upper-primitivity on $\emptyset\neq\Fix{T}\subset\op{int}(\sgens{k}_+).$
\end{remark}

From the above remark, we note that providing an easily verifiable sufficient condition of lower- and upper-primitivity to hold at an arbitrary $u\in\Fix{T}$ is important for obtaining information about the shape of the fixed point set of neural networks. Fortunately, for neural networks that are differentiable at their fixed points, the following sufficient condition exists. We note that, in view of Remark \ref{lu_primitive}, it is sufficient to verify this condition for a particular fixed point $u\in\Fix{T}$ to guarantee that it holds for every other point in $\Fix{T}.$

\begin{lemma} \label{suff_L_U}
Consider an $(A_{0,1,2})$-neural network $T:\op{int}(\sgens{k}_+)\to\op{int}(\sgens{k}_+)$ of the form (\ref{dnn}) along with its continuous extension (see Remark~\ref{extension}). Let $T$ have a fixed point $u\in\op{int}(\sgens{k}_+)$, and let $T$ be differentiable at $u\in\op{int}(\sgens{k}_+).$ Then $T$ is both lower- and upper-primitive at $u$ if the Jacobian matrix of $T$ evaluated at $u$ is primitive.
\end{lemma}
\proof Since $T$ is differentiable at $u\in\op{int}(\sgens{k}_+)$, it is Fréchet differentiable at $u\in\op{int}(\sgens{k}_+)$ with $DT_u(v)=J_T(u)v$ for $v\in\sgens{k}$, where $DT_u\colon\sgens{k}\to\sgens{k}$ is the Fréchet derivative of $T$ at $u$, and $J_T(u)$ is the Jacobian matrix of $T$ evaluated at $u.$ Then the assertion of the lemma follows from \cite[Lemma~6.5.7]{Lemmens2012} specialized to a Fréchet-differentiable mapping at $u.$ $\blacksquare$  

\begin{remark} \label{suff_L_U_remaako}
With notation as in Lemma \ref{suff_L_U}, the Jacobian $J_T(x)$ of $T$ evaluated at $x\in\op{int}(\sgens{k}_+)$ is nonnegative owing to the monotonicity of $T.$ Moreover, $J_T(x)$ is primitive in the sense of Definition \ref{primitivo} if and only if $\exists p\in\mathbb{N}$ s.t. $J_T^p(x)$ is positive, i.e., all of its entries are strictly positive.
\end{remark}

We close Section \ref{wsinn} by anticipating that, later in Fact \ref{conv} in Section \ref{wsinn_arb}, we establish convergence of the fixed point iteration of any  $(A_{0,1})$-neural network (and, hence, of any $(A_{0,1,2})$-neural network) to its fixed point.

\section{Fixed points of $(A_{0,1})$-neural networks} \label{wsinn_arb}
The results of Section \ref{wsinn} provide a largely complete answer to the problem of determining fixed point(s) of $(A_{0,1,2})$-neural networks satisfying Assumption \ref{A012} in Section \ref{wsinn}. We now proceed to relax the assumptions imposed on the neural networks studied so far. In particular, we extend the concepts of nonnegativity and monotonicity to mappings defined on the whole space as follows: 
\begin{definition}
A continuous mapping $f\colon\sgens{s}\to\sgens{p}$ is said to be
\begin{enumerate} [label=(\subscript{A'}{{\arabic*}}), start=0]
       \item \emph{globally nonnegative} if
       \begin{equation} 
       \forall x\in\sgens{s}\quad f(x)\in\sgens{p}_+,
       \end{equation}
       \item \emph{globally monotonic} if
       \begin{equation} 
       \forall x,\tilde{x}\in\sgens{s}\quad x\leq\tilde{x}\implies f(x)\leq f(\tilde{x}),
       \end{equation}
\end{enumerate}
where $u \le v \Leftrightarrow v-u\in\sgens{s}_+$ for $u,v\in\sgens{s}.$ 
\end{definition}

We note that, with $\zeta\in\real$, the sigmoid: $\zeta\mapsto\frac{1}{1+\exp{(-\zeta)}}$ and ReLU: $\zeta\mapsto 0$ for $\zeta<0$ and $\zeta\mapsto \zeta$ for $\zeta\geq 0$ are activation functions that are in $(A_{0',1'}).$

Below, we replace Assumption~\ref{A012} with the following assumptions, which, in particular, lift the restriction of nonnegative biases in all layers and enlarge the class of possible activation functions.

\begin{assumption} \label{A01}
For all layers $i=1,\dots,n$:
\begin{enumerate}
\item The weight matrices $W_i$ have nonnegative coefficients.
\item Let $S\in 2^{\{1\dots,n\}}$, where $2^{\{1\dots,n\}}$ denotes the power set of $\{1,\dots,n\}$, be the set of layers with activation functions $\sigma_s$ that belong to the class $(A_{0',1'})$ (e.g., layers constructed with the ReLU or sigmoid scalar activation functions), where $s\in S$. The biases $b_s$ are allowed to be negative.
\item Define $P\df\{1,\dots,n\}\backslash S$. Then, the activation functions $\sigma_p$ for $p\in P$ are required to be in the class $(A_{0,1}).$ The biases $b_p$ are assumed to be nonnegative.

\end{enumerate}
\end{assumption}

To emphasize that Assumption~\ref{A01} covers well-known activation functions that have not been considered in our analysis until this moment, we list the following activation functions, which  are in $(A_{0,1})$ but not in $(A_{0,1,2})$ for $\xi\in\mathbb{R}_+$:
\begin{enumerate}
	\item[(L3)] scalar $(A_{0,1})$-activation functions:
	\begin{itemize}
		\item (Mish)
		$\xi\mapsto \xi\tanh(\log(1+\exp(\xi)))$
		\item (Swish)
		$\xi\mapsto \frac{\xi}{1+\exp{(-\xi)}}$
		\item (GELU)
		$\xi\mapsto \xi\Phi(\xi)$,
		where $\Phi(\xi)\df\mathbb{P}(X\leq\xi),\quad X\sim\mathcal{N}(0,1).$
	\end{itemize}
\end{enumerate}

\begin{corollary} \label{A01_corr}
If a neural network $T$ of the form (\ref{dnn}) satisfies Assumption \ref{A01}, then $T$ is an $(A_{0,1})$-neural network.
\end{corollary}
\proof If $S\neq\emptyset$, then nonnegativity of layers $T_S$ for $s\in S$ is implied by the assumption that the activation functions $\sigma_s$ are nonnegative on the whole $\sgens{k_s}.$ To establish monotonicity of $T_s$ for $s\in S$, let $x,\tilde{x}\in\sgens{k_{s-1}}_+$ such that $x\leq\tilde{x}.$ Then, by Assumption \ref{A01}, point 1, $W_sx\leq W_s\tilde{x}$, and by Assumption \ref{A01}, point 2, $\sigma_s(W_sx+b)\leq\sigma_s(W_s\tilde{x}+b)$ for any $b\in\sgens{k_s}.$ Similarly, if $P\neq\emptyset$, then $T_p$ for $p\in P$ is an $(A_{0,1})$-layer from Lemma \ref{construction}, point 1, followed by an application of Lemma \ref{fact_combs}, point 2(a). The fact that $T$ is an $(A_{0,1})$-neural network now follows by application of Lemma \ref{fact_combs}, point 2(a), across layers. $\blacksquare$ 

As the following examples show, we may leverage the results of Section \ref{wsinn} together with Fact \ref{Renato_fact} from Section \ref{renato_ext} below to establish the existence of fixed points of $(A_{0,1})$-neural networks.


\begin{example} \label{ex_swishmish_1} Assume that $T_2\colon\sgens{k}_+\to\sgens{k}_+$ satisfies Assumption \ref{A01} with $S=\{1,\dots,n\}$ and $P=\emptyset$ and such that the $(A_{0',1'})$-activation functions $\sigma_s$ are in $(A_{0,1,2})$ if their domains are restricted to $\sgens{k_s}_+$, e.g., composed of ReLU or sigmoid activation functions for $s\in S.$ Define $L\colon\sgens{k}_+\to\sgens{k}_+$ by replacing the negative biases on layers $S$ with arbitrary nonnegative values. Such an $L$ is an $(A_{0,1,2})$-neural network by Lemma \ref{construction}, point 3, followed by Lemma \ref{construction}, point 6, and we have that $\forall x\in\sgens{k}_+\ T_2(x)\leq L(x).$ We may now use Corollary \ref{nonemptylo} and Fact \ref{Renato_fact} to conclude that, if $\rho(L_\infty)<1$, then $\Fix{T_2}\neq\emptyset.$
\end{example}

\begin{example} \label{ex_swishmish_2}
Assume that $T_2\colon\sgens{k}_+\to\sgens{k}_+$ satisfies Assumption \ref{A01} with $S=\emptyset$ and $P=\{1,\dots,n\}.$ Construct a neural network $L\colon\sgens{k}_+\to\sgens{k}_+$ from $T_2$ by replacing the activation functions in layers $P$ with pointwise upper-bounding $(A_{0,1,2})$-activation functions, e.g., by replacing the Swish, Mish, or GELU activation functions with the ReLU function. Then, analogously as in the previous example, the neural network $L$ is an $(A_{0,1,2})$-neural network by Lemma \ref{construction}, point 3, and Lemma \ref{construction}, point 6, and such that $\forall x\in\sgens{k}_+\ T_2(x)\leq L(x).$ We can now use Corollary \ref{nonemptylo} and Fact \ref{Renato_fact} to conclude that $\Fix{T_2}\neq\emptyset$ if $\rho(L_\infty)<1$.
\end{example}

If $T_2$ satisfies Assumption \ref{A01} with $\emptyset\neq S\subsetneq\{1,\dots,n\}$, then the previous two examples can be combined to obtain a pointwise upper-bounding network $L$ that can be used to infer the existence of fixed points of $T_2$ if $\rho(L_\infty)<1.$ 

We close Section \ref{wsinn_arb} with the following fact yielding convergence of the fixed point iteration of an $(A_{0,1})$-neural network to its fixed point. This fact is well-known, and it can be obtained, for example, by following the proof of \cite[Proposition 3]{Cavalcante2019}. We remark that, although the statement of \cite[Proposition 3]{Cavalcante2019} considers mappings in class $(A_{0,1,2})$, the proof uses only the assumptions of continuity and monotonicity of the mappings under consideration.
\begin{fact}[\cite{Cavalcante2019}] \label{conv}
Let $T$ be an $(A_{0,1})$-neural network of the form (\ref{dnn}) with $\Fix{T}\neq\emptyset$. Then, there exists $x^\star\in \Fix{T}$ such that, for all $y\in\Fix{T}$, we have $x^\star\leq y.$ Furthermore, this least fixed point $x^\star$ is obtained as the limit of the fixed point iteration $x_{n+1}\df T(x_n)$ of $T$ with $x_1=0.$
\end{fact}

\section{Fixed points of $(A_{0})$-neural networks} \label{renato_ext}
In Section~\ref{renato_ext}, we further relax the assumptions made so far to include in our analysis neural networks that are only nonnegative. More specifically, we replace Assumption~\ref{A012} with the following:

\begin{assumption} \label{A0}
The final $n$-th layer of the neural network has an $(A_0')$-activation function $\sigma_n.$ All other layers are only required to be constructed with continuous activation functions.
\end{assumption}

We emphasize that Assumption \ref{A0} imposes no restrictions on the weight matrices or biases. They can have arbitrary real coefficients in any layer. As an immediate consequence of Assumption~\ref{A0}, we have the following result.

\begin{corollary} \label{A0_cor}
If a neural network $T$ of the form (\ref{dnn}) satisfies Assumption \ref{A0}, then $T$ is an $(A_{0})$-neural network.
\end{corollary}

We close the analytical derivations of this paper with a simple result establishing the existence of a fixed point of an $(A_{0})$-neural network. 

\begin{fact} \label{Renato_fact}
Let $L\colon\sgens{k}_+\to\sgens{k}_+$ be an $(A_{0,1})$-neural network. Assume that $\Fix{L}\neq\emptyset.$ Let $T_2:\sgens{k}_+\to\sgens{k}_+$ be an $(A_0)$-neural network  such that $\forall x\in\sgens{k}_+\ T_2(x)\leq L(x).$ Then $\Fix{T_2}\neq\emptyset.$
\end{fact}
\proof Define $F\df\{x\in\sgens{k}_+\colon\ x\leq x^\star\}$, where $x^\star\in\Fix{L}.$ We note that $F$ is convex and compact in the standard Euclidean metric space. Therefore,
\begin{equation}
\forall x\in F\quad \sgens{k}_+ \ni T_2(x)\leq L(x)\leq L(x^\star)=x^\star,
\end{equation}
which proves that $T_2(x)\in F$ for all $x\in F.$ As a result, $T_{2|{F}}$ is a continuous self-mapping on $F$, and, by Brouwer's fixed point theorem, it possesses a fixed point in $F.$ $\blacksquare$

\begin{remark} \label{A0_remark_1}
From Fact \ref{Renato_fact}, we conclude that, in order to establish the existence of fixed points of an $(A_0)$-neural network $T_2$ of the form (\ref{dnn}), it suffices to upper-bound $T_2$ by an $(A_{0,1})$-neural network $L$, which could be, for example, one satisfying Assumption \ref{A01} (see Corollary~\ref{A01_corr}). Then, the existence of fixed points of both $T_2$ and $L$ can be established by upper-bounding $L$ (for example, with the approaches in Examples \ref{ex_swishmish_1} and \ref{ex_swishmish_2}) by an $A_{0,1,2}$-neural network $T$ and by using the results of Section \ref{wsinn}.
\end{remark}

\begin{remark} \label{A0_remark_2}
The existence of a fixed point of an $(A_0)$-neural network can be established using Fact \ref{Renato_fact}, but the fixed point iteration of such a network does not necessarily converge to its fixed point. 
\end{remark}

Below, we discuss applications of the analytical results derived in this paper.

\section{Applications} \label{app}
\subsection{Autoencoders} \label{ne}
For all experiments in this section, we trained a two-layered autoencoder with input and output layers of dimension $k=784$ and hidden layer with $k_1=200$ neurons.\footnote{The source code is available at the following link: \url{https://github.com/mateuszgabor/nn_networks}.} The networks were trained for 30 epochs using the ADAM optimization algorithm with a learning rate of 0.005. The batch size was set to 64, and, as a loss function, the mean squared error was chosen. To enforce nonnegativity of the weights and biases, the negative values were clipped to zero after each iteration of the ADAM algorithm. The experiments were performed on both the entire MNIST dataset and a subset containing only the digit ``zero,'' which we refer to as the ZERO dataset. We considered various constraints on the weights and biases, so, to simplify notation, we use the following acronyms in the text that follows:
\begin{itemize}
\item NN denotes an autoencoder with nonnegative weights and nonnegative biases;
\item PN denotes an autoencoder with positive weights and nonnegative biases;
\item NR denotes an autoencoder with nonnegative weights and real-valued biases; and
\item RR denotes an autoencoder with real-valued weights and real-valued biases.
\end{itemize}

\subsubsection{Sigmoid NN} \label{sec:sigmoid}
\begin{figure}[h!]
    \centering
    \subfloat[]{%
        \includegraphics[width=0.25\linewidth]{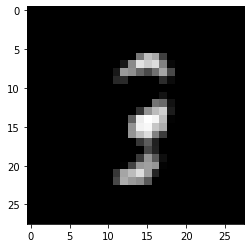}%
        }%
    \subfloat[]{%
        \includegraphics[width=0.25\linewidth]{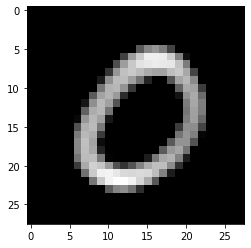}%
        }%
    \caption{The unique fixed points of an autoencoder with sigmoid activation functions and nonnegative weights and biases on (a) the full MNIST dataset and (b) the ZERO dataset.}
    \label{fig:sigmoid}
\end{figure}

In this setup, the autoencoder $T$ uses the sigmoid activation functions in both layers. Thus, according to Corollary~\ref{ex1}, point \ref{ex1_3}, $T$ is an $(A_{0,1,2,3})$-neural network. The sigmoid activation function is in the list (L1) in Remark \ref{activ}, and, hence, from Proposition \ref{sr}, point \ref{sr_2}, and Corollary \ref{sr_cor}, the spectral radius of $T$ is $\rho(T_{\infty}) = 0.$ Thus, from Corollary \ref{uniqulo}, $T$ has a unique and positive fixed point. Moreover, it follows from Fact \ref{conv} that this fixed point can be computed with the fixed point iteration of $T$ starting at $x_1=0.$ The fixed points obtained on the MNIST and ZERO datasets are presented in Figure \ref{fig:sigmoid}.

\subsubsection{Tanh NN} \label{sec:tanh}
\begin{figure}[h!]
    \centering
    \subfloat[]{
        \includegraphics[width=0.9\linewidth]{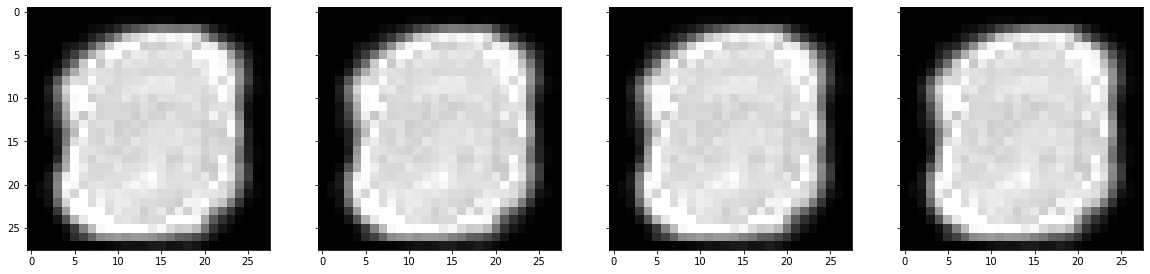}%
        }
        \\
     \subfloat[]{
        \includegraphics[width=0.7\linewidth]{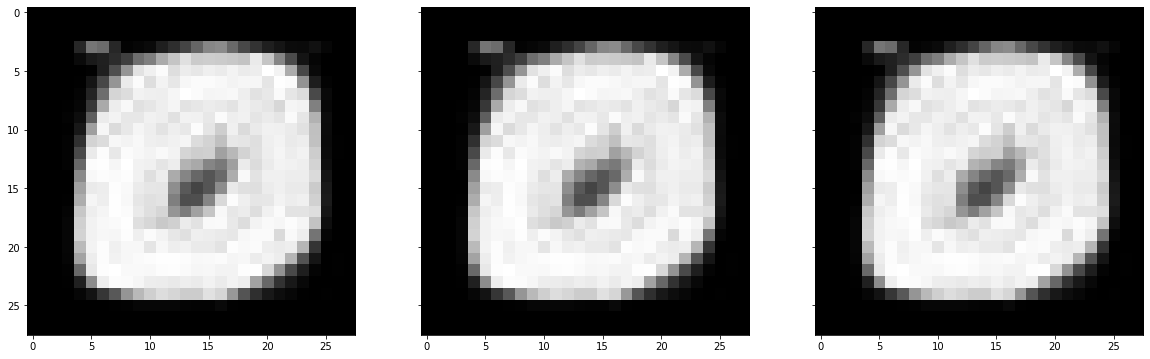}%
        }
        \caption{Sample fixed points of an autoencoder with hyperbolic tangent activation functions and with nonnegative weights and biases on (a) the full MNIST dataset and (b) the ZERO dataset.}
        \label{fig:tanh}
\end{figure}

In this setup, the autoencoder $T$ uses hyperbolic tangent activation functions in both layers. Thus, according to Corollary~\ref{ex1}, point \ref{ex1_1}, $T$ is an $(A_{0,1,2})$-neural network. The hyperbolic tangent activation function is in the list (L1) in Remark \ref{activ}, hence, from Proposition \ref{sr}, point~\ref{sr_2}, and Corollary \ref{sr_cor}, the spectral radius of $T$ is $\rho(T_{\infty}) = 0.$ Thus, from Corollary~\ref{nonemptylo}, $T$ has at least one fixed point. Moreover, it follows from Fact \ref{conv} that the least fixed point of $T$ can be computed using the fixed point iteration of $T$ starting at $x_1=0.$ We have also checked that the other fixed points of $T$ can be obtained for different positive starting points of the fixed point iteration of $T.$ However, the sufficient condition of lower- and upper-primitivity of $T$ at its fixed point given in Lemma \ref{suff_L_U} was not satisfied by $T$, thus the shape of the fixed point set of $T$ could not be determined. Numerically, the fixed points obtained in the MNIST and ZERO datasets are shown in Figure \ref{fig:tanh}. For the considered case, we have empirical evidence that the shape of the fixed point set is probably an interval, but there is no theory to prove formally this claim at this moment, and this topic should be investigated in future studies.

\subsubsection{Tanh PN}
\label{sec:tanh_primitive}
\begin{figure}[h!]
    \centering
    \subfloat[]{
        \includegraphics[width=0.7\linewidth]{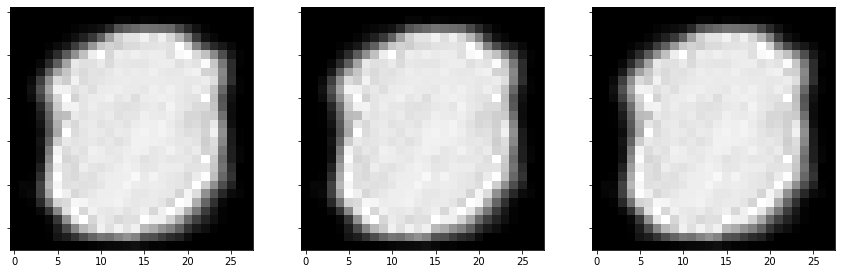}%
        }
        \\
     \subfloat[]{
        \includegraphics[width=0.5\linewidth]{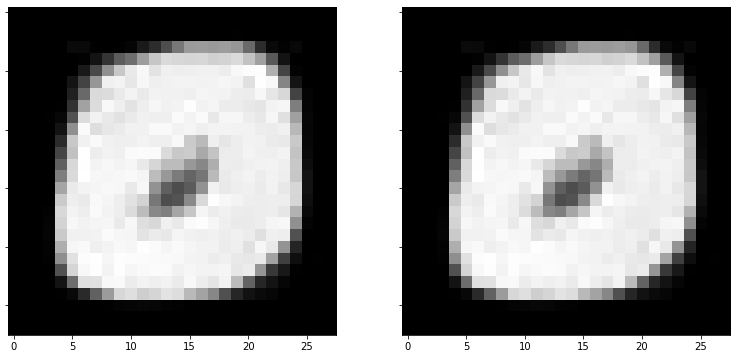}%
        }
        \caption{Sample fixed points of an autoencoder with a primitive Jacobian,  positive weights, nonnegative biases, and hyperbolic tangent activation functions on (a) the full MNIST dataset and (b) the ZERO dataset.}
        \label{fig:primitive_jacobian}
\end{figure}

In this setup, we  modified slightly the autoencoder $T$ in Section \ref{sec:tanh} by clipping the negative coefficients of the weights in training to 0.001 instead of 0. The resulting autoencoder $T$ has a primitive Jacobian matrix evaluated at a fixed point of $T$. Hence, by Lemma \ref{suff_L_U} and Remark \ref{lu_primitive}, such an autoencoder is both lower- and upper-primitive on the whole $\Fix{T}.$ Thus, from Proposition \ref{p3}, we deduce that $\Fix{T}$ is an interval and that the fixed point iteration of $T$ converges to $u\in\Fix{T}$ for any positive starting point $x_1\gg 0.$ Figure \ref{fig:primitive_jacobian} depicts fixed points of $T$ obtained for various positive starting points of the fixed point iteration. We note that the fixed points are very similar to the fixed points obtained for the autoencoder in Section \ref{sec:tanh}.

\subsubsection{ReLU spectral NN}
\label{sec:relu}
\begin{figure}[h!]
    \centering
    \subfloat[]{
        \includegraphics[width=1\linewidth]{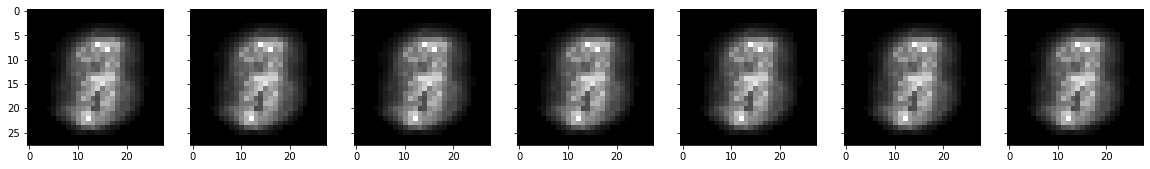}%
        }
        \\
    \subfloat[]{
    \includegraphics[width=0.7\linewidth]{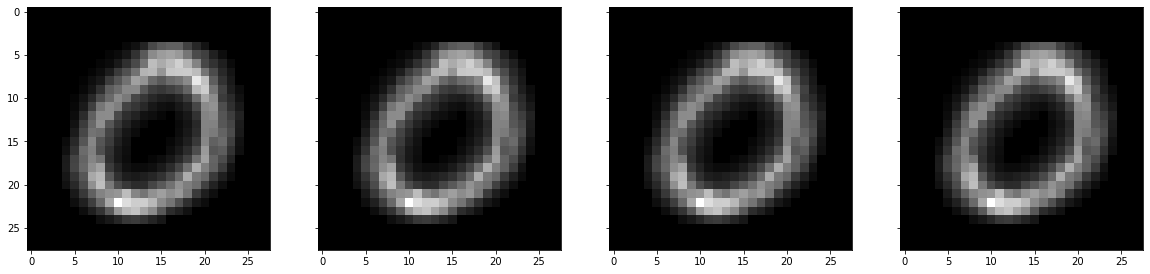}%
    }
    \caption{Sample fixed points of an autoencoder with nonnegative weights and biases and with ReLU activation functions using spectral normalization on (a) the MNIST dataset and (b) the ZERO  dataset.}
    \label{fig:relu_spectral}
\end{figure}

In this setup, the autoencoder $T$ used the ReLU activation function in both layers. Thus, according to Corollary~\ref{ex1}, point \ref{ex1_1}, $T$ is an $(A_{0,1,2})$-neural network. The ReLU activation function is in the list (L2) in Remark \ref{activ}. Hence, from Proposition \ref{sr}, point \ref{sr_1}, and Corollary~\ref{sr_cor}, the spectral radius of $T$ is $\rho(T_\infty)=\max\{\lambda\in\mathbb{R}_+:\ \prod_{i=n}^1W_ix=\lambda x\}.$ After training, we obtained a neural network $T$ with $\rho(T_{\infty}) = 1.01 > 1$ for both the MNIST and ZERO datasets. In such a case, according to \cite[Proposition 2]{Cavalcante2019}, the fixed-point of $T$ may only exist on the boundary of the cone. Indeed, we have numerical evidence that, in this case, the only fixed point of $T$ is the zero vector. Moreover, the obtained autoencoder had a high test loss compared to other setups. Therefore, to obtain a nondegenerate fixed point set of $T$, we enforced the spectral radius of the autoencoder to be less than~1 using spectral normalization \cite{Miyato2018}, which normalizes all weight values by the largest singular value of the weight matrix. Indeed, the spectral radius obtained by the autoencoder $T$ trained with spectral normalization on the MNIST dataset is $\rho(T_{\infty}) = 0.93$, and, on the ZERO dataset, we have $\rho(T_{\infty}) = 0.91$. Thus, from Corollary \ref{nonemptylo}, $T$ has at least one fixed point. Moreover, it follows from Fact \ref{conv} that the least fixed point of $T$ can be computed with the fixed point iteration of $T$ starting at $x_1=0.$ We have also checked that the other fixed points of $T$ can be obtained for different positive starting points of the fixed point iteration of $T.$ However, similarly to the Tanh NN autoencoder in Section \ref{sec:tanh} above, the sufficient condition of lower- and upper-primitivity of $T$ at its fixed point given in Lemma \ref{suff_L_U} was not satisfied by $T$, thus the shape of the fixed point set of $T$ could not be determined. Numerically, the fixed points obtained in the MNIST and ZERO datasets are shown in Figure~\ref{fig:relu_spectral}. For this case under consideration, the shape of the fixed point set is probably an interval, but, as already mentioned, at this moment there is no theory to prove formally this claim.

\subsubsection{ReLU spectral PN} \label{sec:relu_primitive}
\begin{figure}[h!]
    \centering
    \subfloat[]{
        \includegraphics[width=0.5\linewidth]{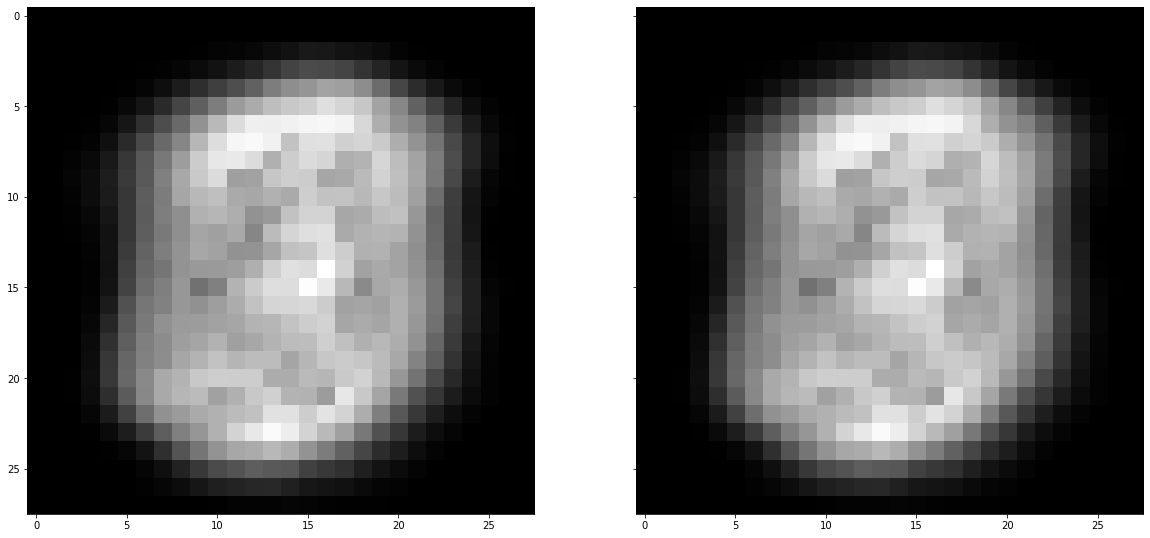}%
        }
        \\
     \subfloat[]{
        \includegraphics[width=0.7\linewidth]{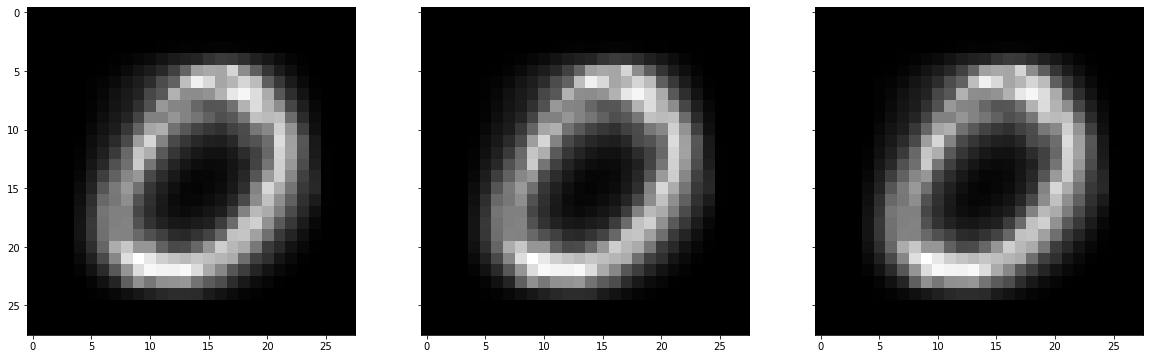}%
        }
        \caption{Sample fixed points of an autoencoder with primitive Jacobian, positive weights, nonnegative biases, and ReLU activation functions on (a) the MNIST dataset and (b) the ZERO dataset.}
        \label{fig:primitive_jacobian_relu}
\end{figure}

In this setup, we modified slightly the autoencoder $T$ in Section \ref{sec:relu} by clipping the negative coefficients of the weights during training to 0.001 instead of 0. The resulting autoencoder $T$ has a primitive Jacobian matrix evaluated at a fixed point of $T$. Hence, by Lemma \ref{suff_L_U} and Remark~\ref{lu_primitive}, such an autoencoder is both lower- and upper-primitive on the whole $\Fix{T}.$ Thus, from Proposition \ref{p3}, we conclude that $\Fix{T}$ is an interval and that the fixed point iteration of $T$ converges to some $u\in\Fix{T}$ for any positive starting point $x_1\gg 0.$ Figure~\ref{fig:primitive_jacobian_relu} presents sample fixed points of $T$ obtained for various positive starting points of the fixed point iteration.

\subsubsection{Tanh + Swish NN}
\label{sec:tanh_swish}
\begin{figure}[h!]
    \centering
    \subfloat[]{%
        \includegraphics[width=0.9\linewidth]{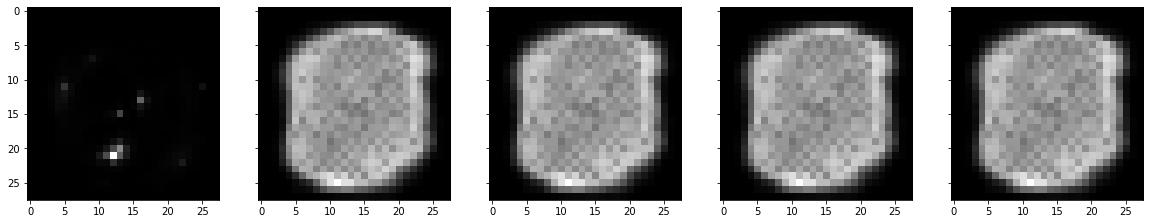}%
        }%
        \\
     \subfloat[]{%
        \includegraphics[width=0.6\linewidth]{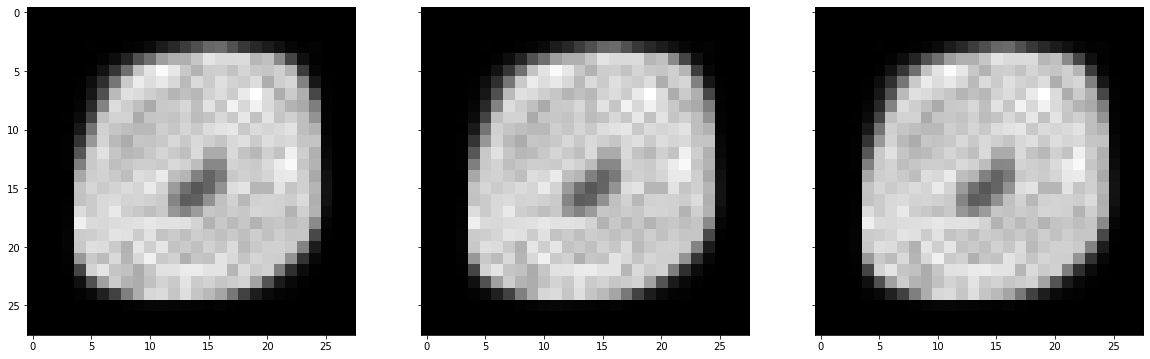}%
        }%
    \caption{Sample fixed points of an autoencoder with nonnegative weights and biases and with the hyperbolic tangent and Swish activation functions on (a) the MNIST dataset and (b) the ZERO dataset.}
    \label{fig:tanh_swish}
\end{figure}

In this setup, the autoencoder $T_2$ used the hyperbolic tangent activation functions in the first layer and the Swish activation functions in the second layer. Such an autoencoder satisfies Assumption \ref{A01} for $S=\emptyset$ and $P=\{1,2\}$, and thus it is an $(A_{0,1})$-neural network by Corollary \ref{A01_corr}. Then, following Example \ref{ex_swishmish_2}, this autoencoder is pointwise upper-bounded by an autoencoder $L$ constructed by replacing the Swish activation functions with the ReLU activation functions in the second layer, and we verify that $L$ is an $(A_{0,1,2})$-neural network. Therefore, using Proposition \ref{sr}, point \ref{sr_2}, and Corollary \ref{sr_cor}, we obtain that $\rho(L_{\infty}) = 0$, and, consequently, from Corollary \ref{nonemptylo} we know that $\Fix{L}\neq\emptyset.$ Hence, from Fact \ref{Renato_fact}, we conclude that the autoencoder $T_2$ has at least one fixed point, and, from Fact \ref{conv}, we obtain that the least fixed point of $T_2$ is the limit of the sequence constructed via the fixed point iteration of $T_2$ starting at $x_1=0.$ We have also checked that the other fixed points of $T_2$ can be obtained for different positive starting points of the fixed point iteration of $T_2.$ The sample fixed points obtained for $T_2$ are depicted in Figure \ref{fig:tanh_swish}.
 
\subsubsection{ReLU + Sigmoid NR} \label{sec:relu_sigmoid_nr}
\begin{figure}[h!]
    \centering
    \subfloat[]{%
        \includegraphics[width=0.5\linewidth]{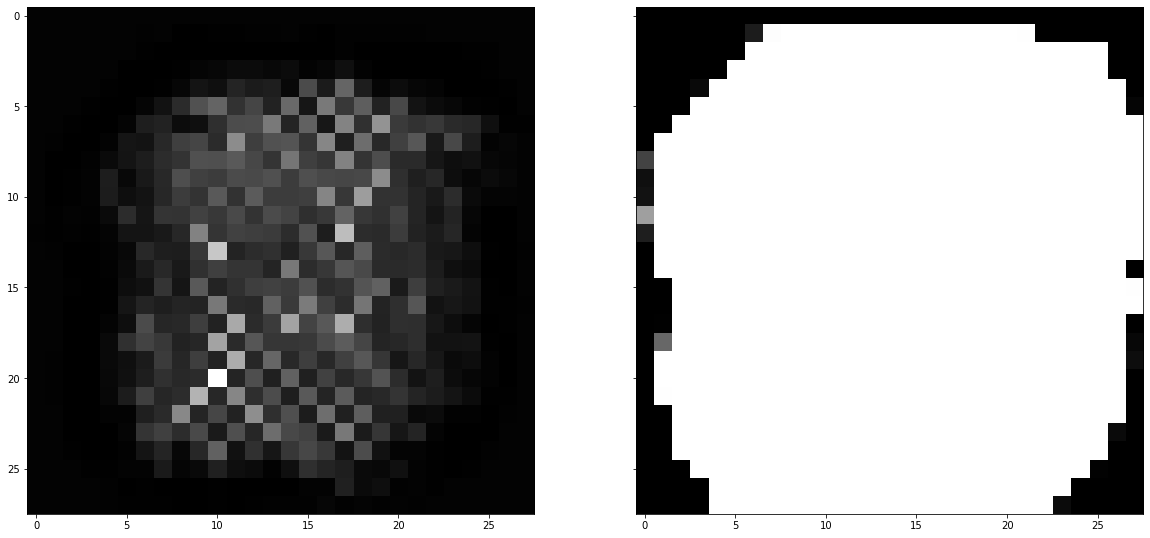}%
        }%
        \\
    \subfloat[]{%
        \includegraphics[width=0.9\linewidth]{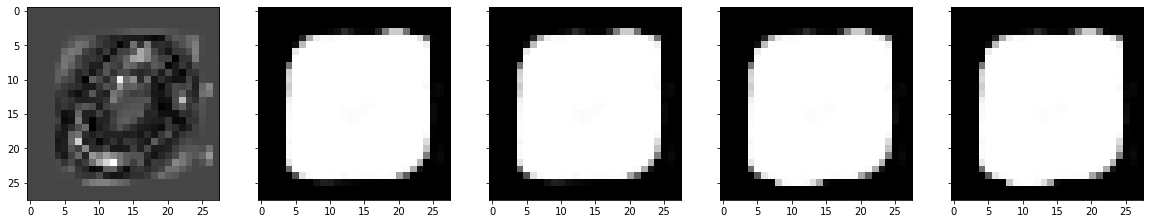}%
        }%
        \caption{Sample fixed points of an autoencoder with only nonnegative weights and with the ReLU and sigmoid activation functions on (a) the MNIST dataset and (b) the ZERO dataset.}
        \label{fig:nn_weights_sigmoid_relu}
\end{figure}

In this setup, the autoencoder $T_2$ had nonnegative weights and real-valued biases, and it used ReLU activation functions in the first layer and sigmoid activation functions in the second layer. Such an autoencoder satisfies Assumption \ref{A01} for $S=\{1,2\}$ and $P=\emptyset$, and thus it is an $(A_{0,1})$-neural network by Corollary \ref{A01_corr}. Then, following Example \ref{ex_swishmish_1}, such an autoencoder is pointwise upper-bounded by the autoencoder $L$ with the same structure but with negative biases replaced with zeros. By doing so, $L$ is an $(A_{0,1,2})$-neural network. Therefore, using Proposition \ref{sr}, point \ref{sr_2}, and Corollary \ref{sr_cor}, we verify that $\rho(L_{\infty}) = 0$. Consequently, from Corollary \ref{nonemptylo} we deduce that $\Fix{L}\neq\emptyset.$ Hence, from Fact \ref{Renato_fact} we conclude that the autoencoder $T_2$ has at least one fixed point, and, from Fact \ref{conv}, we know that the least fixed point of $T_2$ can be computed using the fixed point iteration of $T_2$ starting at $x_1=0.$ We have also checked that the other fixed points of $T_2$ can be computed for different positive starting points of the fixed point iteration of $T_2.$ The sample fixed points obtained for $T_2$ are depicted in Figure \ref{fig:nn_weights_sigmoid_relu}.

\subsubsection{ReLU + Sigmoid RR} \label{sec:relu_sigmoid_rr}
\begin{figure}[h!]
    \centering
    \includegraphics[scale=0.37]{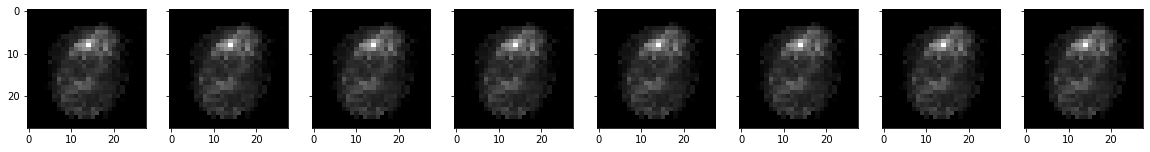}
    \caption{The sample fixed points of an autoencoder with real-valued weights and biases on the MNIST dataset.}
    \label{fig:real_autoencoder}
\end{figure}

In this setup, the autoencoder $T_2$ had real-valued weights and biases, and it used ReLU activation functions in the first layer and sigmoid activation functions in the second layer. Thus, the autoencoder satisfies Assumption \ref{A0}, and, hence, by Corollary \ref{A0_cor}, $T_2$ is an $(A_0)$-neural network. By replacing negative weights and biases with zeros, we obtain from Corollary~\ref{ex1}, point~\ref{ex1_3}, an $(A_{0,1,2,3})$-neural network $L$ that upper-bounds $T_2$. From Proposition \ref{sr}, point \ref{sr_2}, and Corollary \ref{sr_cor}, we have that $\rho(L_{\infty}) = 0$, and, from Corollary \ref{uniqulo}, we conclude that $L$ has a unique and positive fixed point. We note that $L$ is in particular an $(A_{0,1})$-neural network, which follows from the relations $(A_{0,1,2,3})\subset (A_{0,1,2})\subset(A_{0,1}).$ Therefore, using Fact \ref{Renato_fact} and Remark \ref{A0_remark_1}, we conclude that $\Fix{T_2}\neq\emptyset.$ However, as stated in Remark \ref{A0_remark_2}, the fixed point iteration of $T_2$ does not necessarily converge to its fixed point. Indeed, using the fixed point iteration of $T_2$ for various starting points, we obtained fixed points for the MNIST dataset, but not for the ZERO dataset. The sample fixed points obtained on the MNIST dataset are depicted in Figure~\ref{fig:real_autoencoder}.  

\begin{table}[h!]
\centering
\caption{Empirical mean of the test loss for different configurations of the autoencoders. The test loss is averaged over 10 runs of the autoencoder training. The deviations shown in the table indicate the standard error of the mean, extending one standard error above and below the empirical mean.}
\label{tab:results}
\begin{tabular}{lc}
\toprule
Configuration     & Test loss   \\ \midrule
Sigmoid NN \ref{sec:sigmoid}        & $0.2308\pm 2.3\cdot 10^{-6}$  \\
Tanh NN \ref{sec:tanh}        & $0.0063\pm3.9\cdot 10^{-6} $ \\
Tanh PN  \ref{sec:tanh_primitive}        & $0.0246\pm 5.6\cdot 10^{-5}$  \\
ReLU spectral NN \ref{sec:relu} & $0.0061\pm 1.1\cdot 10^{-5}$  \\
ReLU spectral PN \ref{sec:relu_primitive}  & $0.0059\pm 1.3\cdot 10^{-5}$  \\
Tanh + Swish NN \ref{sec:tanh_swish}  & $0.0041\pm 2.5\cdot 10^{-6}$  \\
ReLU + Sigmoid NR \ref{sec:relu_sigmoid_nr} & $0.0031\pm 4.7\cdot 10^{-5}$  \\
ReLU + Sigmoid RR \ref{sec:relu_sigmoid_rr} & $0.0042\pm 7.7\cdot 10^{-5}$ \\ \bottomrule
\end{tabular}

\centering
\caption{Empirical mean for the spectral radii, products of spectral norms of weight matrices, and spectral norms of product of weight matrices for different configurations of autoencoders.  The deviations shown in the table indicate the standard error of the mean, extending one standard error above and below the empirical mean. All values have been computed by considering 10 runs of the autoencoder training.}
\label{tab:results2}
\begin{tabular}{lccc}
\toprule
Configuration      & $\rho(T_\infty)$ & $\|W_1\| \cdot \|W_2\|$ & $\|W_2 \cdot W_1\|$ \\ \midrule
Sigmoid NN \ref{sec:sigmoid}          & 0.00 & $6.26 \pm 0.07$ & $5.23 \pm 0.03$ \\
Tanh NN \ref{sec:tanh}         & 0.00 & $373.9 \pm 3.8$ & $95.2 \pm 0.6$ \\
Tanh PN  \ref{sec:tanh_primitive}      & 0.00 & $35.5 \pm 0.6$ & $31.9 \pm 0.5$ \\
ReLU spectral NN \ref{sec:relu} & $0.988 \pm 3.7\cdot 10^{-4}$ & $1.066 \pm 4.7 \cdot 10^{-4}$ & $0.999 \pm 4.3 \cdot 10^{-4}$ \\
ReLU spectral PN \ref{sec:relu_primitive}   & $0.987 \pm 7.7\cdot 10^{-5}$ & $1.001 \pm 3.5 \cdot 10^{-5} $ & $0.999 \pm 2.9 \cdot 10^{-5} $\\
Tanh + Swish NN \ref{sec:tanh_swish}  & \NA & $15.8 \pm 0.09$ & $8.96 \pm 0.04$ \\
ReLU + Sigmoid NR \ref{sec:relu_sigmoid_nr}   & \NA & $123.3 \pm 0.7$ & 
$100.7 \pm 0.6$ \\
ReLU + Sigmoid RR \ref{sec:relu_sigmoid_rr} & \NA & $271.5 \pm 0.1$ & $214.7 \pm 1.0$ \\ \bottomrule
\end{tabular}
\end{table}

\subsubsection{Connections between the existence of fixed points, the nonlinear spectral radius, and the spectral norm of the weight matrices}
\label{sect.fixedpointspecradius}
The fixed points obtained in the above simulations on the MNIST dataset resemble a noisy average of input classes, which is visually very different from any reasonable input signal (i.e., handwritten digits from 0 to 9). In scenarios of this type, fast convergence of the autoencoder iterations to the set of fixed points is undesirable because the output of the autoencoder can be seen as the first iteration of a fixed point algorithm. Fast convergence of the autoencoders leads to significant differences between the autoencoder input (the handwritten digits) and the output. To optimize the performance of nonnegative autoencoders, a natural strategy is to ensure slow convergence to fixed points for samples of the expected input classes. We discuss this promising but challenging research line in more detail in Section \ref{slo}. However, to provide precise mathematical statements that motivate this discussion, we need a clear understanding of some key properties: the nonlinear spectral radius and the weight matrices of the trained neural networks.  To this end, we show in Table \ref{tab:results} the performance of the autoencoders considered in this section, and in Table \ref{tab:results2} we show key characteristics of these autoencoders with respect to spectral norms of weight matrices and the nonlinear spectral radius. The important aspect to highlight is that previous studies deriving sufficient conditions for the existence of fixed points (and convergence of the fixed point iteration to the fixed points) would require the spectral norm of the product of weight matrices to be strictly less than one, which is a condition often violated in the experiments depicted in Table \ref{tab:results2}. However, we have verified numerically the existence of fixed points regardless of the values taken by these weight matrices, and these numerical results are consistent with the theory discussed in, for example, Section \ref{existence}. We have also verified that, for the autoencoders for which the concept of nonlinear spectral radius is well defined, the best performing autoencoders have spectral radius close to one. This aspect will be crucial for the discussion later in Section \ref{slo}.

\subsection{Deep equilibrium models}
As discussed in the Introduction, fixed point analysis plays a pivotal role in the development of deep equilibrium models. In these models, we typically need to guarantee existence and uniqueness of the fixed point, in addition to establishing convergence of the fixed point iteration. In our recent study in \cite{Gabor2024}, we leverage the theoretical framework laid out in the preceding sections to achieve these objectives. More specifically, in that study, we use the results related to $(A_{0,1,2,3})$-mappings to construct a deep equilibrium model that is guaranteed to have a unique fixed point. Furthermore, we also establish geometric convergence of fixed point iteration based of the results in \cite{Piotrowski2022}. We refer readers to \cite{Gabor2024} for details.

\section{Future research directions} \label{frd}
Current studies indicate that obtaining simple and general results about the shape of the fixed point set of generic $(A_{0,1})$-neural networks can be difficult. As an example, a very recent result \cite{Reich2023} shows that, even for the simple case of $(A_1)$ (continuous and monotonic) self-mappings defined on an interval $[a,b]$, i.e., nondecreasing functions $f\colon[a,b]\to[a,b]$, the problem of determining explicitly $\Fix{f}$ is a highly non-trivial task. In particular, \cite[Theorem 2.2]{Reich2023} asserts that, for any integer $n\geq 2$, there exists an open and everywhere dense set of such functions for which the cardinality of $\Fix{f}$ is lower bounded by $n$. Furthermore, any uniqueness conditions on fixed points of $(A_{0,1,2})$-mappings, beyond those presented in this paper, must clearly preclude simple cases such as the identity mapping $Id\colon\sgens{k}_+\to\sgens{k}_+$, which is the simplest example of an $(A_{0,1,2})$-mapping with an uncountable number of fixed points ($\Fix{Id}=\sgens{k}_+$).

Two additional important questions, especially from an application perspective, are the following:
\begin{enumerate}
	\item[(i)] determine the sets of approximate fixed points, that is, find $x\in\sgens{k}_+$ such that $T(x)\approx x$ (in some sense), where $T$ is the nonnegative neural network under consideration; and
	\item[(ii)] lift the restriction of nonnegative inputs and outputs.
\end{enumerate}

Below we discuss potential research directions toward these goals.
 
\subsection{Approximate fixed points}

At first glance, the imposition of a nonnegativity constraint on neural networks, as explored in this study and inspired by biological mechanisms, might seem to significantly impair their performance. We might expect, for instance, that this constraint could dramatically hinder the ability of neural networks to reconstruct input data perfectly. Nevertheless, we argue that achieving perfect signal reconstruction is unnecessary for a wide array of applications. To support this viewpoint, we refer to the analogy of human cognition, where the brain is understood to process sensory information to a level considered ``adequate,'' rather than aiming for ``perfection.'' This viewpoint is consistent with established computational neuroscience principles \cite{Doya2007,Arbib2016,Poggio2016}.

 Using this analogy, we posit that autoencoders subjected to nonnegativity constraints can, with proper training, process input data ``sufficiently well'' in some applications of practical interest. This assertion is supported by empirical evidence \cite{lemme2012online,ayinde2018deep}, and our discussions in previous sections link this practical aspect with theoretical insights using simple autoencoders on the MNIST dataset. In particular, as briefly mentioned in Section~\ref{sect.fixedpointspecradius}, viewing the output of an autoencoder as the first iteration of a fixed point algorithm offers a natural strategy for developing autoencoders that closely emulate an ideal encoder-decoder model, which compresses and reconstructs perfectly input signals. The nuances of this strategy are explored in the next two subsections.

\subsubsection{Slow convergence to fixed points} \label{slo}

Let us consider the input-output relation $x\mapsto T(x)$ of a nonnegative autoencoder $T$ as the first iteration of the fixed point algorithm $x_{n+1}\df T^n(x)$ for $n\in\mathbb{N}.$ From this fixed point theory viewpoint, an autoencoder $T$ is expected to provide good reconstruction performance ($x\approx T(x)$) if the fixed point algorithm $x_{n+1}=T^n(x)$ generates slowly converging or diverging sequences. Building on the theoretical foundations laid out in prior sections, we now introduce simple, analytically justified heuristics designed to fulfill this objective.

For the special case of a positive concave autoencoder $T$,\footnote{For example, autoencoders $T$ of the form (\ref{dnn}) that satisfies $T(0)\gg 0$ and that uses activation functions constructed with continuous scalar concave activation functions from lists (L1)-(L2) in Remark \ref{activ}. The Sigmoid NN autoencoder considered in Section \ref{sec:sigmoid} is an example of a positive concave neural network. See also Corollary \ref{cor.uniqueness}.} which belongs to the subclass of $(A_{0,1,2,3})$-autoencoders, the results in \cite[Remark 8]{Piotrowski2022} and \cite[Proposition 5]{cavalcante2018spectral} derive a contraction factor $c\in[0,1)$ that indicates the rate of geometric convergence of the fixed point iteration of $T$ to its unique fixed point in Euclidean spaces. The closer $c$ is to one, the slower the convergence speed of the fixed point algorithm is expected to be, and, hence, the better reconstruction performance we may expect from the autoencoder. While a contraction factor $c$ close to the value one does not provide us with formal guarantees of small progress at the first iteration, we illustrate via a simple numerical experiment that designing autoencoders experiencing potentially slow convergence or divergence in their fixed point iterations is a useful strategy to enhance their performance.

\begin{figure}[h!]
    \centering    \includegraphics[scale=0.6]{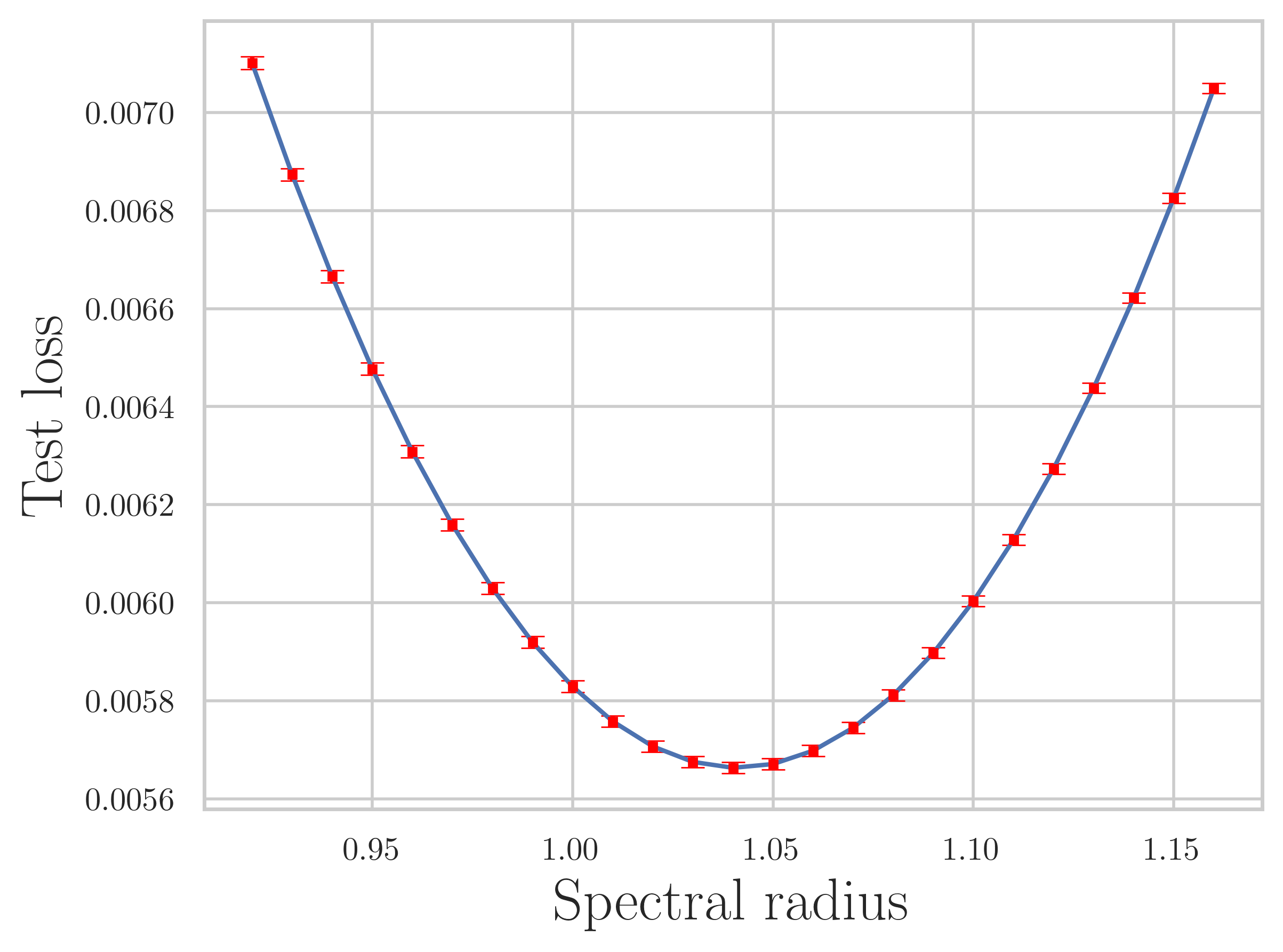}
    \caption{Empirical mean of the test loss as a function of the spectral radius for the ReLU spectral PN autoencoder described in Section \ref{sec:relu_primitive} on the MNIST dataset. The test loss is averaged over 10 runs of the autoencoder training. The error bars indicate the standard error of the mean, extending one standard error above and below the empirical mean.}
    \label{fig:scaling} 
\end{figure}

In more detail, for positive concave autoencoders, the contraction factor $c$ is lower bounded by the spectral radius $\rho(T_\infty)$, and $\rho(T_\infty)<1$ if and only if the autoencoder is $c$-Lipschitz continuous with $c<1$ in a neighborhood of the fixed point with respect to Thompson's metric described in Definition~\ref{tmetric} in the Appendix \cite{Piotrowski2022}.  If $\rho(T_\infty)>0$, it follows from (\ref{fix2}) that we can modify the autoencoder $T$ to have its spectral radius to an arbitrary nonnegative value $u$ by simply rescaling the weights of any layer by $u/\rho(T_\infty)$, so, based on the above arguments, we should set $u$ to a value close to one to improve the performance of autoencoders.  This idea is similar to spectral normalization of neural network layers, but our motivation for normalization is uniquely grounded in fixed point theory, providing further theoretical justifications even in nonlinear settings. To demonstrate that this principle can also be exploited by autoencoders that are not necessarily positive and concave, we conducted an experiment with the trained $(A_{0,1,2})$-ReLU spectral PN autoencoder $T$ from Section \ref{sec:relu_primitive}. We rescaled the weights of the first layer of $T$ to obtain 25 instances of the autoencoder $T$ with spectral radii $[0.92,0.93,0.94,\dots,1.2].$ The results are shown in Figure \ref{fig:scaling}, and we verify that the test loss indeed achieves its minimum value if the spectral radius is close to one. More precisely, if $1-\epsilon<c<1$ for some sufficiently small $\epsilon>0$, we may expect slowly converging fixed point iterations. On the other hand, if $1\le c < 1+\epsilon$, Figure \ref{fig:scaling} suggests that autoencoders with slowly diverging fixed point iterations are also able to produce good approximate fixed points. 

The above considerations constitute only a sample starting point of a much deeper analysis that is necessary to establish approximate fixed points of more general nonnegative neural networks. As another example of a possible direction towards this goal, we note that, for the ReLU spectral PN autoencoder considered above, if $W_1W_2$ is diagonalizable and the input classes are (approximately) within the subspace spanned by the eigenvectors of $W_1W_2$ corresponding to eigenvalues approximately equal to one, we should observe the ``slow-convergence'' behaviour described above, and, consequently, good reconstruction performance of the autoencoder.

\subsubsection{Lipschitz continuity in normed vector spaces}

Many of the networks we have considered in the previous sections are in fact $c$-Lipschitz continuous with $c\le 1$ with respect to Thompson's metric defined in the Appendix. However, we now show that, in normed vector spaces -- which are arguably the spaces of most practical significance for engineers -- providing bounds on the Lipschitz constant is not likely to succeed because the best Lipschitz constant can be prohibitively large in Euclidean settings. This insight provides us with information about the output variability in response to minor input modifications in Euclidean settings. Importantly, it also also provides us with a means of evaluating the robustness of neural networks against adversarial attacks. 



More precisely, the result in \cite[Section~2]{Combettes2020a} shows that all activation functions from lists (L1)-(L2) in Remark \ref{activ} are nonexpansive in a Hilbert space. Thus, from \cite[Proposition 4.3]{Combettes2020b}, we deduce that neural networks satisfying the conditions in Definitions \ref{nn}-\ref{sep_activ}, and, in particular, those considered in Sections \ref{wsinn}-\ref{renato_ext}, are Lipschitz continuous, and the corresponding Lipschitz constant $\theta$ in the standard Euclidean space can be bounded by the spectral norm $\|\cdot \|$ of the weight matrices $W_1, \ldots, W_n$, or, more precisely:
\begin{equation}
\|\prod_{i=n}^1W_i\|\leq\theta\leq\prod_{i=1}^n\|W_i\|.
\end{equation}

The results presented in Table \ref{tab:results2} at the end of Section \ref{ne} show that, for $n=2$, the lower bound of the spectral norm $\|\prod_{i=n}^1W_i\|$ (achievable by neural networks with nonnegative weights \cite[Section~5.3]{Combettes2020b}) is too large to obtain meaningful bounds for $\|T(x)-x\|$ around its fixed point(s) $x^\star\in\Fix{T}$, where the vector norm $\|\cdot\|$ is the standard Euclidean norm. Therefore, a more subtle argument, perhaps based on local Lipschitz continuity for specific input classes, should be considered for exploiting the continuity of $T$ with the aim of determining all inputs $x$ satisfying $T(x)\approx x$, or, equivalently, $\|T(x)- x\|\approx 0$.

\subsection{Extending the current results beyond nonnegativity}
This study mostly focuses on monotonic neural networks with nonnegative inputs and outputs, but a well-known pre- and post-processing step enables us to extend the above results to a particular class of monotonic neural networks with inputs and outputs possibly taking values in the whole real line. More precisely, let $T:\op{int}(\sgens{k}_+)\to\op{int}(\sgens{k}_+)$ be a neural network with a structure that can be analyzed with the theory in this study. To train and operate a neural network of this type with inputs and outputs taking values in $\sgens{k}$, we can proceed as follows. Let $L:\op{int}(\sgens{k}_+)\to \sgens{k}:x\mapsto[\log(x_1),\ldots,\log(x_k)]$ and $E:\sgens{k}\to\op{int}(\sgens{k}_+):x\mapsto[\exp(x_1),\ldots,\exp(x_k)]$ denote, respectively, the coordinate-wise log and exponential mappings. Now, consider the neural network defined by $N:\sgens{k}\to\sgens{k}:x\mapsto L(T(E(x)))$; i.e., we first map an input in $\sgens{k}$ to $\op{int}(\sgens{k}_+)$ with the exponential mapping, process the mapped input with the original positive neural network $T$, and then map the output of $T$ back to $\sgens{k}$ with the log mapping. Some properties of $N$ can be directly deduced from $T$. For example, we verify that $N$ has a fixed point if and only if $T$ also has a fixed point. Furthermore, the uniqueness of the fixed point of $T$ implies the uniqueness of the fixed point of $N$. These are all properties that can be verified with the theory described in the previous sections. From a mathematical perspective, the mapping $L$ is the known isometry between Thompson's metric space $(\Intr, d_T)$ (see Definition \ref{tmetric}) and the normed vector space $(\sgens{k},~\|\cdot\|_\infty)$, and further properties of the neural network $N$ are strongly connected to the theory of topical mappings \cite[Sect.~1.5]{Lemmens2012}. 

\section{Conclusion} \label{conc}
We have shown that the behavior of many neural networks that can be seen as self-mappings with nonnegative weights (as often proposed in the literature) can be rigorously analyzed with the framework of nonlinear Perron-Frobenius theory. We also demonstrated that the concept of the spectral radius of asymptotic mappings is useful for determining whether a neural network has an empty or nonempty fixed point set, and, for an autoencoder, we recall that its fixed point set corresponds to the inputs that can be perfectly reconstructed at the output. Our conditions for the existence of fixed points of nonnegative neural networks are weaker than those previously considered in the literature, which are often based on arguments in convex analysis in Hilbert spaces. 



\acks{The authors are grateful to anonymous reviewers for their
constructive comments which surely promoted the readability of the
revised manuscript. This work was in part supported by the National Centre for Research and Development of Poland (NCBR) under grant EIG CONCERT-JAPAN/04/2021
and by the Federal Ministry of Education and Research of Germany under
grant 01DR21009 and the programme ``Souver\"an. Digital. Vernetzt.''
Joint project 6G-RIC, project identification numbers: 16KISK020K and
16KISK030.}

\appendix

\section{Known results} \label{kru}
\begin{definition}
	\label{tmetric}
We define Thompson's metric by
\begin{equation} 
	\label{eq.thompson} 
	\begin{array}{rl}
d_T:\Intr\times \Intr&\to [0,\infty)\\ 
(x,y)&\mapsto\ln(\max\{M(x,y),M(y,x)\}),
\end{array}
\end{equation}
where \begin{equation*} 
	\begin{array}{rl}
	M: \Intr\times \Intr &\to[0,\infty)\\ 
	(x,y)&\mapsto \inf\{\beta>0~|~x\leq\beta y\}.
	\end{array}
\end{equation*}
\end{definition}
We call the metric space $(\Intr, d_T)$ the \emph{Thompson metric space}.

\begin{proposition}[{\cite[Proposition 1]{Cavalcante2019}}] \label{P1_R2019}
Let $T\colon\sgens{k}_+\to\sgens{k}_+$ be an $(A_{0,1,2})$-mapping. If $\rho(T_{\infty})<1$, then $\Fix{T}\neq\emptyset.$
\end{proposition}

\begin{proposition}[{\cite[Proposition 4]{Cavalcante2019_IEEE_TSP}}] \label{P4_R2019_TSP}
Let $T\colon\sgens{k}_+\to\op{int}(\sgens{k}_+)$ be an $(A_{0,1,2,3})$-mapping. Then $\Fix{T}\neq\emptyset$ if and only if $\rho(T_{\infty})<1.$
\end{proposition}

\begin{proposition}[{\cite[Fact 1(ii)]{Cavalcante2019_IEEE_TSP}}] \label{F1_R2019_TSP}
Let $T\colon\sgens{k}_+\to\sgens{k}_+$ be an $(A_{0,1,2,3})$-mapping. Then $\Fix{T}\neq\emptyset$ is either a singleton or the empty set.
\end{proposition}

\section{Construction of monotonic and weakly scalable mappings} \label{combs}
The following general lemma introduces building blocks to construct $(A_{0,1,2})$-mappings, or even $(A_{0,1,2,3})$-mappings. It extends \cite[Proposition 2.3]{Oshime1992} to mappings that are not necessarily self-mappings.

\begin{lemma} \label{fact_combs}
$\ $
  \begin{enumerate}
  \item Let $f_1\colon\sgens{s}_+\to\sgens{p}_+$ and $f_2\colon\sgens{s}_+\to\sgens{p}_+$ be $(A_0)$-mappings and let $\alpha,\beta\geq 0.$ Consider the following mappings:
\begin{flalign} 
  &x\mapsto (\alpha f_1+\beta f_2)(x),&& \label{combs1a}\\ 
  &x\mapsto\max\{f_1(x),f_2(x)\},&& \label{combs1b}\\ 
  &x\mapsto\min\{f_1(x),f_2(x)\}.&& \label{combs1c}
\end{flalign}
  Then we have:
  \begin{enumerate}
  \item If $f_1,f_2$ are $(A_{0,1})$-mappings, then so are the mappings in~(\ref{combs1a})-(\ref{combs1c}).
  \item If $f_1,f_2$ are $(A_{0,1,2})$-mappings, then so are the mappings in (\ref{combs1a})-(\ref{combs1c}).
  \item If $f_1$ in as $(A_{0,1,2})$-mapping and $f_2$ is an $(A_{0,1,2,3})$-mapping, $\alpha\geq 0$ and $\beta>0$, then the mapping in (\ref{combs1a}) is an $(A_{0,1,2,3})$-mapping.
  \item if $f_1,f_2$ are $(A_{0,1,2,3})$-mappings, then so are the mappings in (\ref{combs1b})-(\ref{combs1c}).
  \end{enumerate}
\item Let $g\colon\sgens{p}_+\to\sgens{r}_+$ be an $(A_0)$-mapping and consider the following mapping:
  \begin{equation} \label{combs2}
    x\mapsto (g\circ f_1)(x).
  \end{equation}
  \begin{enumerate}
  \item If $f_1,g$ are $(A_{0,1})$-mappings, then so is the mapping in~(\ref{combs2}).
  \item If $f_1,g$ are strictly monotonic, then so is the mapping in~(\ref{combs2}).  
  \item If $f_1$ is an $(A_{0,2})$-mapping and $g$ is an $(A_{0,1,2})$-mapping, then the mapping in (\ref{combs2}) is an $(A_{0,2})$-mapping.
  \item If $f_1$ is an $(A_{0,2})$-mapping and $g$ is an $(A_{0,1,2,3})$-mapping, then the mapping in (\ref{combs2}) is an $(A_{0,2,3})$-mapping.
  \item If $f_1$ is an $(A_{0,2,3})$-mapping, and $g$ is a strictly monotonic $(A_{0,2})$-mapping, then the mapping in (\ref{combs2}) is an $(A_{0,2,3})$-mapping.
  \end{enumerate}
  \end{enumerate}
\end{lemma}  
\proof
\proofpart{1}{Properties of mappings in (\ref{combs1a})-(\ref{combs1c})}
\paragraph{1.(a)} Let $x,\tilde{x}\in\sgens{s}_+$ be such that $x\leq\tilde{x}$, let $\alpha,\beta\geq 0$, and let $f_1,f_2$ be $(A_{0,1})$-mappings. Then $\alpha f_1(x)\leq\alpha f_1(\tilde{x})$ and $\beta f_2(x)\leq\beta f_2(\tilde{x}).$ By adding these two inequalities we obtain monotonicity of $x\mapsto (\alpha f_1+\beta f_2)(x).$ Furthermore, we also have $\max\{f_1(x),f_2(x)\}\leq\max\{f_1(\tilde{x}),f_2(\tilde{x})\}$ and $\min\{f_1(x),f_2(x)\}\leq\min\{f_1(\tilde{x}),f_2(\tilde{x})\}.$
\paragraph{1.(b)} Let $x\in\sgens{s}_+$, let $\alpha,\beta\geq 0$, and let $f_1,f_2$ be $(A_{0,1,2})$-mappings. Fix $\rho\geq 1.$ Then $\alpha f_1(\rho x)\leq\alpha\rho f_1(x)$ and $\beta f_2(\rho x)\leq\beta\rho f_2(x).$ By adding these two inequalities we obtain weak scalability of $x\mapsto (\alpha f_1+\beta f_2)(x).$ Furthermore, we have $\max\{f_1(\rho x),f_2(\rho x)\}\leq\max\{\rho f_1(x),\rho f_2(x)\}=\rho\max\{f_1(x),f_2(x)\}$, with the same arguments for weak scalability of $\min\{f_1(\rho x),f_2(\rho x)\}.$
\paragraph{1.(c)} Let $f_1$ be an $(A_{0,1,2})$-mapping and $f_2$ be an $(A_{0,1,2,3})$-mapping, with $\alpha\geq 0$ and $\beta>0$, and let $\rho>1.$ Then $\alpha f_1(\rho x)\leq\alpha\rho f_1(x)$ and $\beta f_2(\rho x)\ll\beta\rho f_2(x).$ By adding these two inequalities we obtain scalability of $x\mapsto (\alpha f_1+\beta f_2)(x).$ 
\paragraph{1.(d)} We have $\max\{f_1(\rho x),f_2(\rho x)\}\ll\max\{\rho f_1(x),\rho f_2(x)\}=\rho\max\{f_1(x),f_2(x)\}$, with the same arguments for scalability of $\min\{f_1(\rho x),f_2(\rho x)\}.$
\proofpart{2}{Properties of mapping in (\ref{combs2})}
\paragraph{2.(a)} Let $x,\tilde{x}\in\sgens{s}_+$ be such that $x\leq\tilde{x}$ and let $f_1,g$ be $(A_{0,1})$-mappings. Then $f_1(x)\leq f_1(\tilde{x})$, and, hence, $g(f_1(x))\leq g(f_1(\tilde{x})).$
\paragraph{2.(b)} As above.
\paragraph{2.(c)} Let $x\in\sgens{s}_+$, and let $f_1$ be an $(A_{0,2})$-mapping and $g$ be an $(A_{0,1,2})$-mapping. Fix $\rho\geq 1.$ From weak scalability of $f_1$, we have $f_1(\rho x)\leq\rho f_1(x)$, and, from monotonicity of $g$, we deduce $g(f_1(\rho x))\leq g(\rho f_1(x)).$ Weak scalability of $g$ yields $g(\rho f_1(x))\leq\rho g(f_1(x)).$
\paragraph{2.(d)} Let $x\in\sgens{s}_+$, and let $f_1$ be an $(A_{0,2})$-mapping and $g$ be an $(A_{0,1,2,3})$-mapping. Fix $\rho>1.$ As above, we have $g(f_1(\rho x))\leq g(\rho f_1(x)).$ Then, from scalability of $g$, it follows that $g(\rho f_1(x))\ll\rho g(f_1(x)).$
\paragraph{2.(e)} Let $x\in\sgens{s}_+$, and let $f_1$ be an $(A_{0,2,3})$-mapping, and $g$ be a strictly monotonic $(A_{0,2})$-mapping. Fix $\rho>1.$ By our assumptions, we have $f_1(\rho x)\ll\rho f_1(x)$, and, from strict monotonicity of $g$, we have $g(f_1(\rho x))\ll g(\rho f_1(x)).$ Weak scalability of $g$ implies $g(\rho f_1(x))\leq\rho g(f_1(x))$, and the proof is completed. $\blacksquare$ 

\section{Proof of Lemma \ref{construction}} \label{pd_construction}
\proof We use the results of Lemma \ref{fact_combs} to prove points \emph{1)-7)} as follows:

\emph{1) } Put $f_1(x)=Wx$ and $g(x)=x+b$, and note that $f_1$ is an $(A_{0,1,2})$-mapping if $W$ is nonnegative, whereas $g$ is a strictly monotonic $(A_{0,2})$-mapping for $b\geq 0.$ Then, from \emph{2(a)} in Lemma \ref{fact_combs}, the affine mapping $y_i=W_ix_{i-1}+b_i$ is monotonic, and from \emph{2(d)} in Lemma \ref{fact_combs}, it is also weakly scalable.

\emph{2) } From point \emph{1) } we have that $y_i=W_ix_{i-1}+b_i$ is an $(A_{0,1,2})$-mapping. To prove scalability, we note that, if $b\gg 0$, then $g(x)=x+b$ is a strictly monotonic $(A_{0,2,3})$-mapping. The desired assertion now follows from \emph{2(e)} in Lemma \ref{fact_combs}.

\emph{3) } Monotonicity of $T_i$ follows from \emph{2(a)} in Lemma \ref{fact_combs} and its weak scalability from \emph{2(d)} in Lemma \ref{fact_combs}. 

\emph{4) } Monotonicity of $T_i$ follows from point \emph{3) } above. Scalability of $T_i$ follows from \emph{2(e)} in Lemma \ref{fact_combs}.

\emph{5) } Monotonicity of $T_i$ follows from point \emph{3) } above. Scalability of $T_i$ follows from \emph{2(f)} in Lemma \ref{fact_combs}.

\emph{6) } Monotonicity and weak scalability of $T$ follows from, respectively, an application of \emph{2(a)} and \emph{2(d)} in Lemma~\ref{fact_combs} to consecutive layers of $T.$

\emph{7) } Monotonicity of $T$ follows from point \emph{6) } above. To prove scalability, verify from \emph{2(e)} in Lemma~\ref{fact_combs} that a subnetwork of $T$ composed of the first $i_0$ layers of $T$ is scalable. Scalability of $T$ now follows  from an application of \emph{2(f)} in Lemma~\ref{fact_combs} to consecutive layers of $T$, starting with the $i_0+1$-st layer. $\blacksquare$

\bibliography{IEEEabrv,references}

\end{document}